# *BrainFusionNet*: a deep learning and XAI model to understand local, global, and sequential features of MRI images for improved brain tumour detection

Md Taimur Ahad[1], Bo Song[2*] and Yan Li[1]

**Abstract**
The noise of Magnetic Resonance Imaging (MRI) poses challenges for Deep Learning (DL) when tumor boundaries are obscured, tumor location and appearance are complex due to overlap between tumor and non-tumor cells, and modality identification is difficult because tumor features vanish in the later layers of the DL. Effective feature extraction from given MRI is a possible solution to overcome this challenge. Therefore, we develop BrainFusionNet that combines Convolutional Neural Networks (CNNs), Vision Transformers (ViT), and Gated Recurrent Units (GRUs) to extract spatial, contextual, and sequential features from MRI images for improved brain tumor classification. Furthermore, explainable AI such as SHAP, LIME, and Grad-CAM are integrated to visualise and highlight image regions that contribute to BrainFusionNet's decision-making process. The proposed BrainFusionNet model is evaluated on two publicly available MRI datasets. K-fold validation suggests 98% accuracy on both datasets. The model was compared with the six state-of-the-art (SOTA) CNNs and transfer learning. Among the SOTA CNNs, DenseNet121 and VGG16 achieved the highest accuracy of 96%. The novelty of BrainFusionNet is that the hybrid model effectively extracts local and global features from MRI images, even in small-scale tumor regions and small tumor sizes. The model has a balanced sequential CNN architecture to capture low-level and deeper-layer features; a customized ViT that captures local features, stabilizes gradient flow, and reduces the risk of vanishing gradients during MRI image training. The CNN and ViT outputs are fed into a GRU for final classification. Furthermore, we analyze pixel intensities to determine whether MRI image quality affects image classification. Our findings are very novel in image interpretation, as we found that the distribution of pixel intensities in MRI images affects DL performance.

*Correspondence:
Bo Song
Bo.Song@unisq.edu.au
[1]School of Mathematics, Physics and Computing, University of Southern Queensland, Toowoomba, QLD 4350, Australia
[2]School of Engineering, University of Southern Queensland, Toowoomba, QLD 4350, Australia

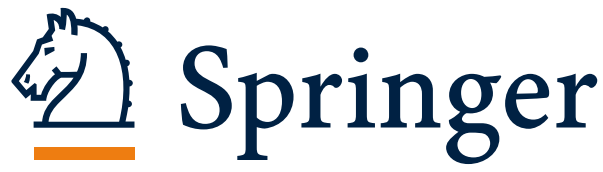

## 1 Introduction

Magnetic Resonance Imaging (MRI) is an effective imaging technique for brain tumor detection. MRI images offer spatial, contextual, and temporal features that are important for tumor detection and classification [8]. However, some challenges exist when Deep Learning (DL) models are used to detect and classify brain tumors. Noise in an MRI image can obscure a tumor, making it difficult to determine its boundaries, shape, and size. Identifying brain tumors in MRI images is also complex due to tumors' hidden locations within the brain, which makes it challenging to differentiate tumors from surrounding tissues. A tumor may appear more visible in a T1-weighted image but less distinguishable in a T2-weighted image [26]. Moreover, tumors, especially in the early stages, may have subtle differences in intensity compared to healthy tissue. Low-grade tumors may not exhibit distinct features in MRI images [45].

Despite the challenges, DL has proven effective methods in detecting brain tumors. The multilayered, hierarchical, and block structure of DL can extract low, mid, and high-level features from MRI images. Whereas standard clinical tumor detection processes require handcrafted feature extraction and manual tumor segmentation before doctors and radiologists classify the tumor, DL provides an end-to-end solution by extracting relevant features and thoroughly analyzing images. Realizing the advantages of hybrid DL, scholars such as Agarwal et al. [3], Kibriya et al. [40] Das and Mohanty [23], Anusha and Kumar [13], Ahmed et al. [6] and Aly et al. [11] applied DL in brain tumor detection.

A hybrid model that combines two or more Convolutional Neural Networks (CNNs) is a practical approach for developing DL systems [53]. The hybrid model improves feature extraction and enhances image classification performance [46]. One main reason for combining two or more CNNs is that each CNN uses a different feature extraction method [27]. For example, sequential CNNs are designed to capture global features, while Vision Transformer (ViT) focuses on local features by analyzing small image patches [25].

The research problem at the core of this study is the difficulty of controlling vanishing features and gradient loss of tumors in MRI images, which hinders the accurate detection of brain tumors using conventional methods. Moreover, a brain tumor has a complex cellular composition, with a mix of malignant tumor cells, stromal cells, and immune cells that interact dynamically and often unpredictably. If the tumor classification is improper, a benign brain tumor can develop into malignant brain cancer [63]. Brain tumors have a complicated pattern of lesion locations. It is difficult to understand the nature of a brain tumor due to its complexity, small scale, and size [47]. Moreover, manual brain tumor detection and classification are time-consuming and error-prone [1]. However, the early detection and accurate classification of brain tumor modalities are essential for a successful and effective treatment plan for brain tumor patients. A research matrix is included in Table 1, which summarizes CNN models and experimental results. A research matrix is included in Table 1, which summarizes CNN models and experimental results.

Despite many studies proposing DL models for detecting and classifying brain tumors, there remain challenges and knowledge gaps in the literature.

1. Scholars have raised the challenge of obtaining tumor location and spatial details in deeper layers of DL due to the small-scale tumor regions compared to the vast difference in area occupancy [20, 44].
2. Accuracy in brain tumor detection is a concern, as lower detection accuracy and large false-positive values narrow the applicability and acceptability of CNNs in brain tumor research [24, 41, 43, 44].
3. Many studies have been conducted using small datasets and limited brain tumor modalities. For instance, the Ranganathan et al. [60] study considered 253 MRI images, and the Aykat [14] dataset also comprised 253 MRI images. In such scenarios, Iqbal et al. [35] advocated for more in-depth and comprehensive research on brain tumor classification.
4. Khan et al. classified CNN networks as spatial exploitation, depth-based, multipath, width-based multiconnection, or feature map exploration [38]. However, there is limited understanding of which CNN architectures can extract more features from MRI images for tumor classification.
5. Rafferty et al. [56, 57] highlighted that in brain tumor detection via CNN models, the tumor image is expected to be visualized to understand the decision-making process of the CNN models. Veeramuthu et al. [68, 69] also reported insufficient interpretability in brain tumor image classification.

To address these gaps, this study conducts three experiments, which make the following contributions and novelties:

1. In this study, a hybrid DL model, *BrainFusionNet*, is developed combining CNN, ViT, and a GRU for improved brain tumor detection and classification. The CNN is used to capture local features, and ViT is employed as a self-attention mechanism to process MRI images as sequences of patches for feature extraction. Finally, the GRU is used to detect and classify the features of brain tumor MRI images. The model was tested on two (2) MRI datasets.

**Table 1** Research matrix

| Author | Model | Accuracy | Contribution |
|---|---|---|---|
| Ibrahim et al. [33] | MobileNetV2 | 99.00% | Contribution of lightweight CNN in real-time brain tumour detection |
| Nahiduzzaman et al. [53] | Lightweight parallel depth-wise separable CNN | 99.35%, 99.30%, and 99.22% | |
| Ishfaq et al. (2025) | Custome CNN, Inception-v4, and EfficientNet-B4 | 97.58%, 99.56%, and 99.76% | |
| Hammad et al. [31] | Lightweight CNN | 99.48% accuracy for binary classification and 96.86% for multiclass classification | |
| Reddy and Dhuli [61] | Lightweight CNN | Classification accuracy of 99.58% and a Dice similarity coefficient (DSC) of 95.7% | |
| Khushi et al. [39] | LeNet, AlexNet, VGG16, VGG19, and ResNet50 | AlexNet achieved the highest accuracy of 98.79% | Improved our understanding of SOTA CNNs' capabilities |
| Hashemzehi et al. [32] | Hybrid CNN-NADE, Alexnet, HCS, SVM | 95%, 96.61%,94.03%, 87.95% | A Hybrid CNN-NADE Model |
| Kibriya et al. (40) | CNN | 97.2% | Create a lightweight CNN architecture |
| Haq et al. [30] | MCNN model | 99.89% | Improved Convolutional Neural Network |
| Ahmed et al. [7] | Efficientnetb0 & SHAP | 99.84% | Improved convolutional neural network (CNN) and Shapley additive explanation (SHAP) |
| Agarwal et al. [3] | Transfer learning VGG16 | 90% | |
| Sun [65] | Customized CNN, ResNet-50, 50 and VGG-16 | 96% | Customized CNN model, ResNet-50 model, and VGG-16 model in the experiments |
| Fuad et al. [29] | Alexnet & GoogleNet | 94.6% & 92% | Compared Alexnet and the GoogleNet architectures on T1-w MRI images |
| Amin et al. [12] | Inceptionv3 | 99.7% | Proposed a deep feature extraction approach |
| Anusha and Kumar [13] | VGG-19 | 98.87% | Improvised classification using deep learning |
| Tataei Sarshar et al. 66] | Resnet-50 | Dice scores of 0.9203, 0.9113, and 0.8726 | Utilizing a feature extraction CNN architecture |
| Zhang et al. [46] | Deep learning | Dice scores of 71%, 88%, and 74% | Segmentation with multiple encoders |

2. Six (6) CNN networks (spatial exploitation (VGG19), depth-based (ResNet152v2), multipath (DenseNet201), width-based multiconnection (ResNext101), feature-map exploitation (SE-ResNet152), and MobileNetv2) are tested on the same dataset to understand which CNN network is more effective in MRI image classification.
3. Explainable AI, such as LIME, SHAP, and Grad-CAM, is integrated with *BrainFusionNet* to understand the model's decision-making process in tumor detection and classification.
4. Finally, the pixel intensities of correct classification, misclassification, true positive, false positive, true negative, and false negative cases were analyzed to understand how pixel intensity impacts detection and classification.

## 2 Related works

Several studies focus on extracting local features through convolutional neural networks (CNNs) or hybrid models. For instance, Kumar et al. [42] used Grey-Level Co-occurrence Matrix (GLCM) for feature extraction followed by a Hybrid Recurrent Neural Network-Bidirectional Long Short-Term Memory (HRNN-BiLSTM) model for classification, achieving an accuracy of 97.8%. Similarly, Yadav et al. [72] applied a probabilistic neural network for feature extraction and achieved 99.46% classification accuracy, showcasing the power of local feature extraction for brain tumor detection. However, these methods face limitations, such as the potential loss of global context as the models focus on local details. Local feature modeling often leads to an incomplete understanding of the global relationships among regions of interest, which is crucial for complex medical imaging tasks such as brain tumor detection.

Recent studies, such as those by Tataei Sarshar et al. [66], have sought to combine local and global feature extraction by employing CNN architectures that integrate ResNet-50 weights. Their method achieved competitive Dice scores for tumor segmentation across multiple modalities. However, a challenge with this approach is that while it captures both local and global features, it may still suffer from insufficient interpretability and spatial detail loss in deeper layers.

In contrast, models like Vision Transformers (ViT), as in Ahmed et al. [6, 9], aim to capture global context through self-attention mechanisms, enabling a more holistic understanding of the entire image. ViT-based methods, such as Ahmed et al.'s [6, 9] hybrid ViT-GRU model, achieve strong performance with 97% in precision, recall, and F1-score. While these methods perform well at capturing global features, they face computational complexity and overfitting, especially as the number of layers increases. Incorporating self-attention mechanisms has become a popular approach in neuroimaging, especially in ViT models. Vamsidhar et al. [67] integrated ResNet101 and Xception with LIME (Local Interpretable Model-Agnostic Explanations) to improve model explainability, achieving an accuracy of 99.67%. The attention mechanism helps the model focus on critical regions of the brain MRI, improving performance but introducing new interpretability challenges—particularly in the absence of clear insights into the attention distribution.

The combination of CNNs and Transformers is an emerging approach to capture both local and global features. Kumar et al. [42] and Vamsidhar et al. [67] have reported promising results with hybrid models that combine CNN-based feature extraction with transformers to understand global context. Ahmed et al. [6, 9] proposed a hybrid ViT-GRU model that not only extracts features via ViT but also models their dependencies using GRUs. The model achieved impressive results with 97% precision, recall, and F1-scores. Despite these advances, challenges remain with model interpretability and feature representation. Hybrid CNN-Transformer models can capture complex patterns but may overfit, especially when not properly regularized. The hybrid nature of such models also introduces increased computational costs and the need for more efficient training methods.

Thus, while attention-based architectures offer significant performance improvements, they often lack clarity in model decision-making and spatial context, leaving room for further improvements in both explainability and spatial accuracy. Afnaan et al. [2] also suggested enhancing transparency in CNN-based classification and improving model performance through advanced optimization strategies. The study applied various dynamic learning rate modifiers and XAI methods to three brain tumor datasets. Among the tested architectures, the enhanced ResNet model consistently outperformed others across all datasets, achieving the highest test accuracy, ranging from 99.36 to 99.65%. Ahmed et al. [6, 9] enhanced the transparency and reliability of their ViT-GRU model by integrating explainable AI (XAI) techniques, making the model's decision-making process more transparent. Despite these advancements, ViT models still struggle to understand feature representations, which limits their ability to generalize across diverse brain tumor datasets. Among other studies, Iftikhar et al. [34] and Bianconi et al. [18] applied CNNs and XAI to MRI evaluation. The proposed algorithm can reduce the impact of low-quality images and missing sequences.

*BrainFusionNet*, our proposed model, addresses these gaps by combining local feature extraction with CNNs and global context modeling with transformers, while also integrating GRUs for relational dependency learning. This hybrid approach is designed to enhance both the accuracy and interpretability of brain tumor classification while addressing key limitations in existing models, including loss of spatial detail and insufficient feature representation. Despite significant advancements, existing methods face a variety of challenges beyond accuracy:

- Loss of spatial detail: As models become deeper or more complex (e.g., hybrid CNN-Transformer), the spatial resolution of features tends to degrade, impacting the fine-grained detection of tumor boundaries.
- Interpretability: While some approaches (e.g., LIME and XAI) attempt to improve model transparency, the deep learning models used in brain tumor detection often operate as "black boxes," making it difficult to understand how the model reaches its decisions.
- Limited understanding of feature representations: Although hybrid models combine multiple approaches, a clear understanding of what features the model focuses on remains largely unexplored. This hinders the ability to generalize models across diverse datasets.

*BrainFusionNet* addresses these gaps by introducing a novel hybrid architecture that combines CNN-based local feature extraction, transformer-based global feature modeling, and GRU for feature relationship learning. This unique integration enhances both local and global feature extraction, improving tumor classification performance. Additionally, our model incorporates techniques that preserve spatial resolution, mitigate overfitting, and enhance interpretability through explainable AI. This makes *BrainFusionNet* a more robust and transparent model compared to existing state-of-the-art methods, particularly for neuroimaging tasks. By explicitly addressing limitations in accuracy, spatial detail loss, and interpretability, *BrainFusionNet* represents a significant step forward in brain tumor classification, positioning it as an advanced solution in brain informatics.

## 3 Research methodology

This section describes the hardware specifications, dataset description, *BrainFusionNet* model development, and training procedure for this study. Figure 1 illustrates the research methodology.

### 3.1 Hardware specification

The experiments were conducted on a Precision 7680 Workstation with a 13th-generation Intel® Core™ i9-13950HX vPro CPU, an NVIDIA® RTX™ 3500 Ada Generation GPU, 32 GB DDR5 RAM, a 1 TB SSD and Windows 11 Pro operating system. Python (version 3.9) is a programming language because it supports TensorFlow-GPU, SHAP, and LIME generation.

### 3.2 Dataset description

In the experiments, two sets of brain MRI image datasets were utilized. Dataset A consists of three classes: Glioma, Pituitary, and Meningioma [22]. Dataset B is a combination of Figshare, SARTAJ, and Br35H. The 'No tumor' class images were sourced from the Br35H dataset.

To ensure consistency and avoid data leakage, the MRI data were split at the patient level. Each patient's MRI scans were kept together as a single unit, ensuring that no scans from different patients were mixed during the dataset split. The data distribution is shown in Fig. 2.

### 3.3 Image augmentation

Image augmentation has been reported as a feasible and practical approach for enhancing the robustness of CNN models. This study uses image transformations, including scaling, rotation, and contrast adjustment, to enhance the training datasets and improve CNN performance across various complex tasks (see Fig. 3). Specifically, in the context of brain tumor detection, the technique allows models to handle variability in tumor presentation, including differences in size, shape, and location.

## 4 Model development and results

The following sections describe the model development and results.

### 4.1 BrainFusionNet model development

The proposed *BrainFusionNet* consists of three different CNN layers: a lightweight CNN, a customised ViT, and a GRU-based classifier. The graphical view of *BrainFusionNet* is presented in Fig. 4.

#### *4.1.1 CNN layer*

The lightweight method consists of six convolutional layers arranged in three pairs, with progressively increasing filter counts: 32, then 64, and finally 128. The input image size is $112 \times 112 \times 3$ (see Fig. 5A-B). Each convolutional layer employs a $3 \times 3$ kernel, ReLU activation, and 'same' padding to preserve spatial dimensions. To downsample the feature maps and introduce a degree of translational invariance, max pooling layers (pool size 2 × 2) are used. The feature maps from the final convolutional layer are flattened and then transformed by a dense layer with 512 neurons and ReLU activation. The output of a convolutional layer method can be described in Eq. (1). $W$ denotes the kernel weights, $b$ Is the bias, and $ReLU$ Adds nonlinearity.

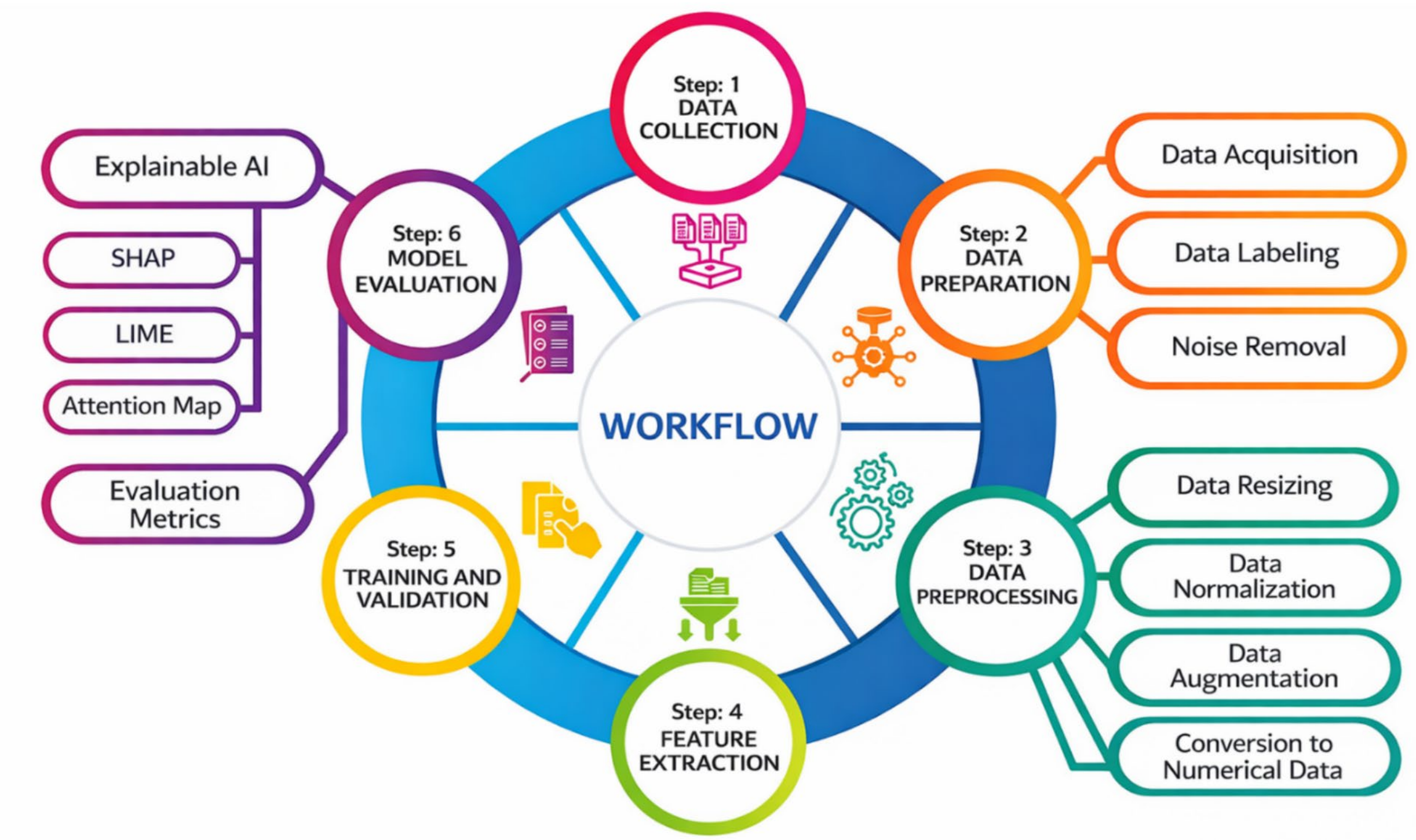

**Fig. 1** The research methodology adopted in this study

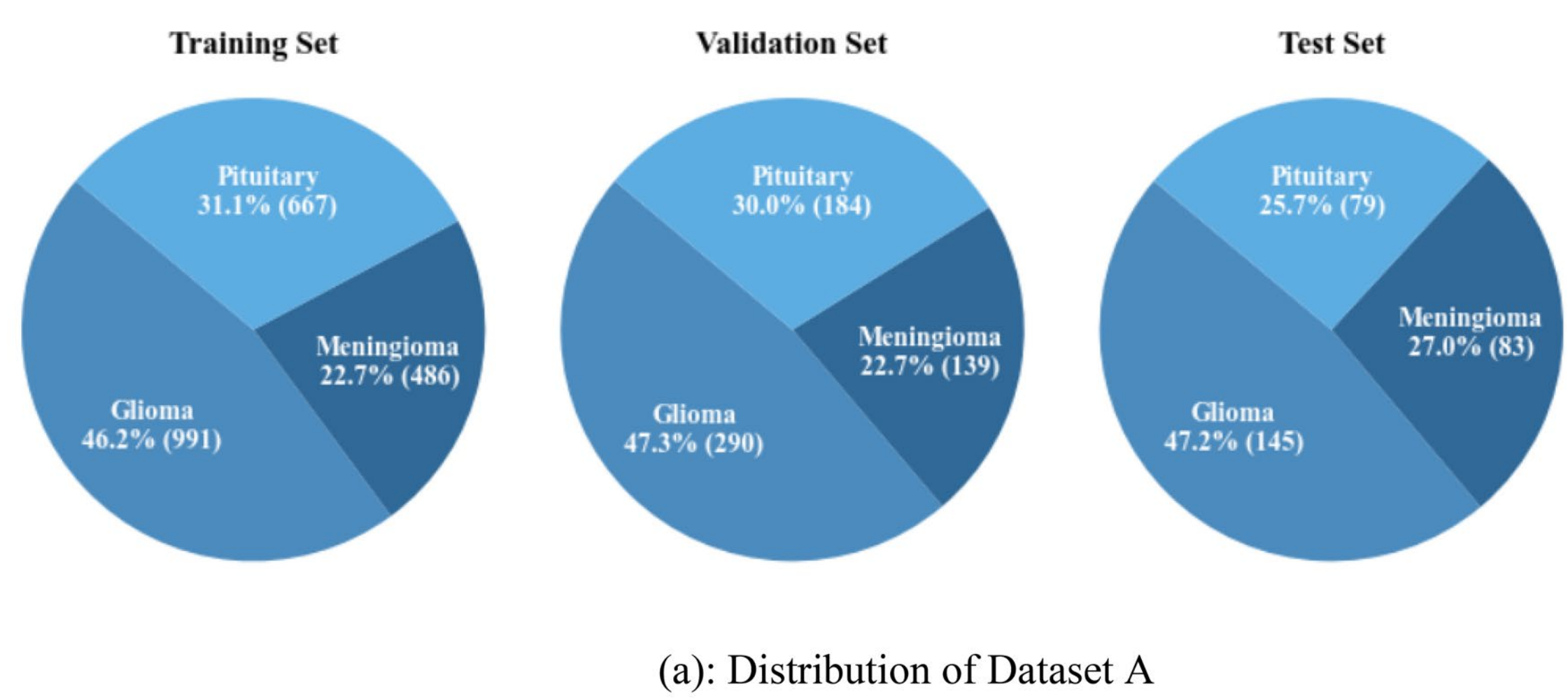

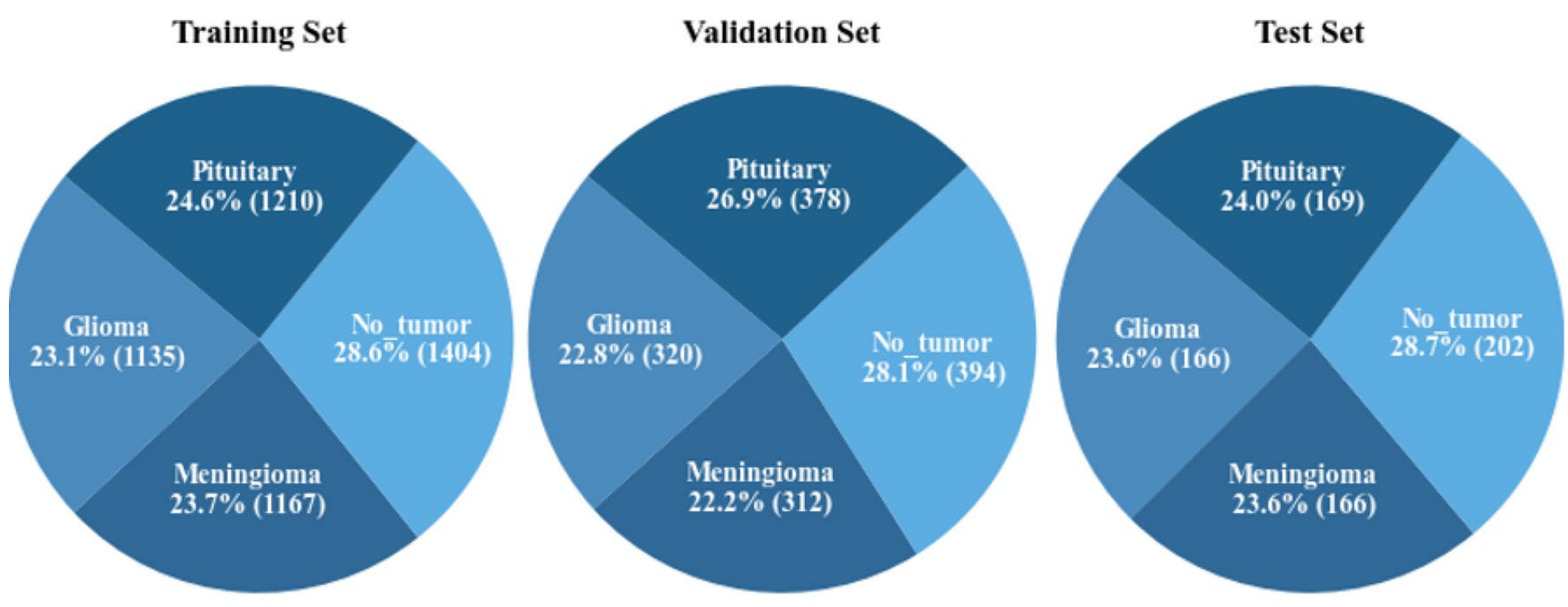

**Fig. 2** **a** Distribution of Dataset A. **b** Distribution of Dataset B

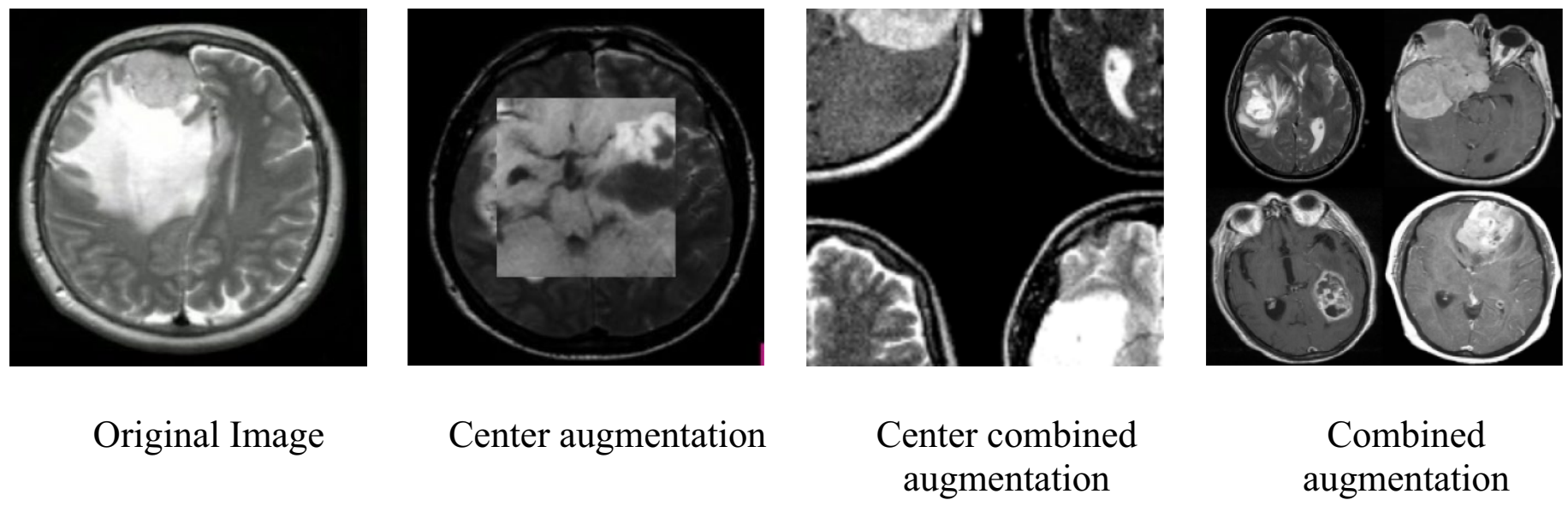

**Fig. 3** Images after augmentation

$$f_{CNN} = ReLU\left(W_{dense} \cdot Flatten\left(X\right) + b_{dense}\right) \quad (1)$$

#### *4.1.2 ViT*

The ViT is utilized to capture global features from the MRI image (see Fig. 6). The input image is partitioned into patches via a convolutional projection, with a kernel size and stride equal to a predetermined patch size (PATCH_SIZE). Each patch is then linearly embedded into a dimensional feature vector, which is mathematically presented in Eq. (2):

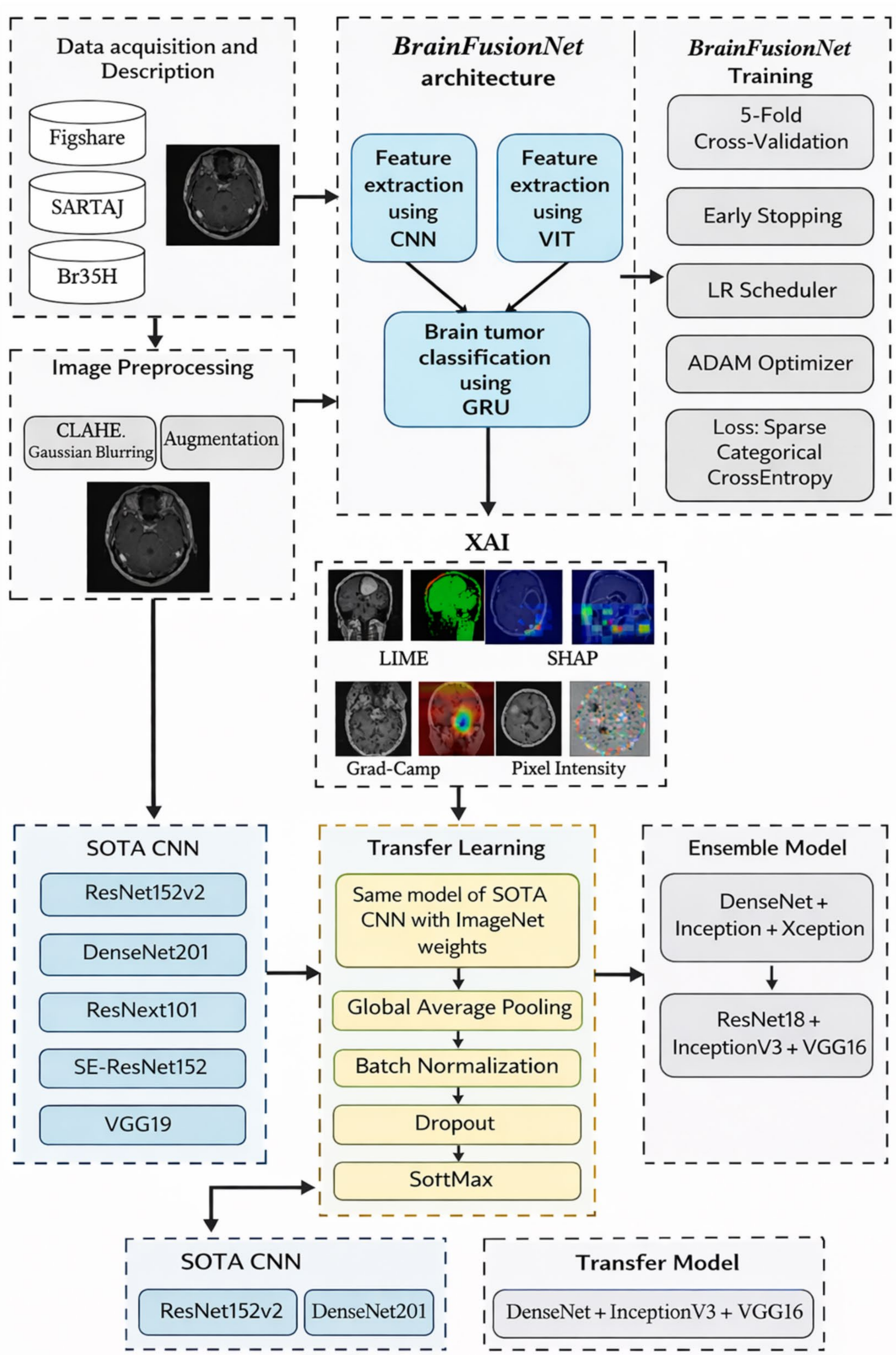

**Fig. 4** Graphical view of BrainFusionNet

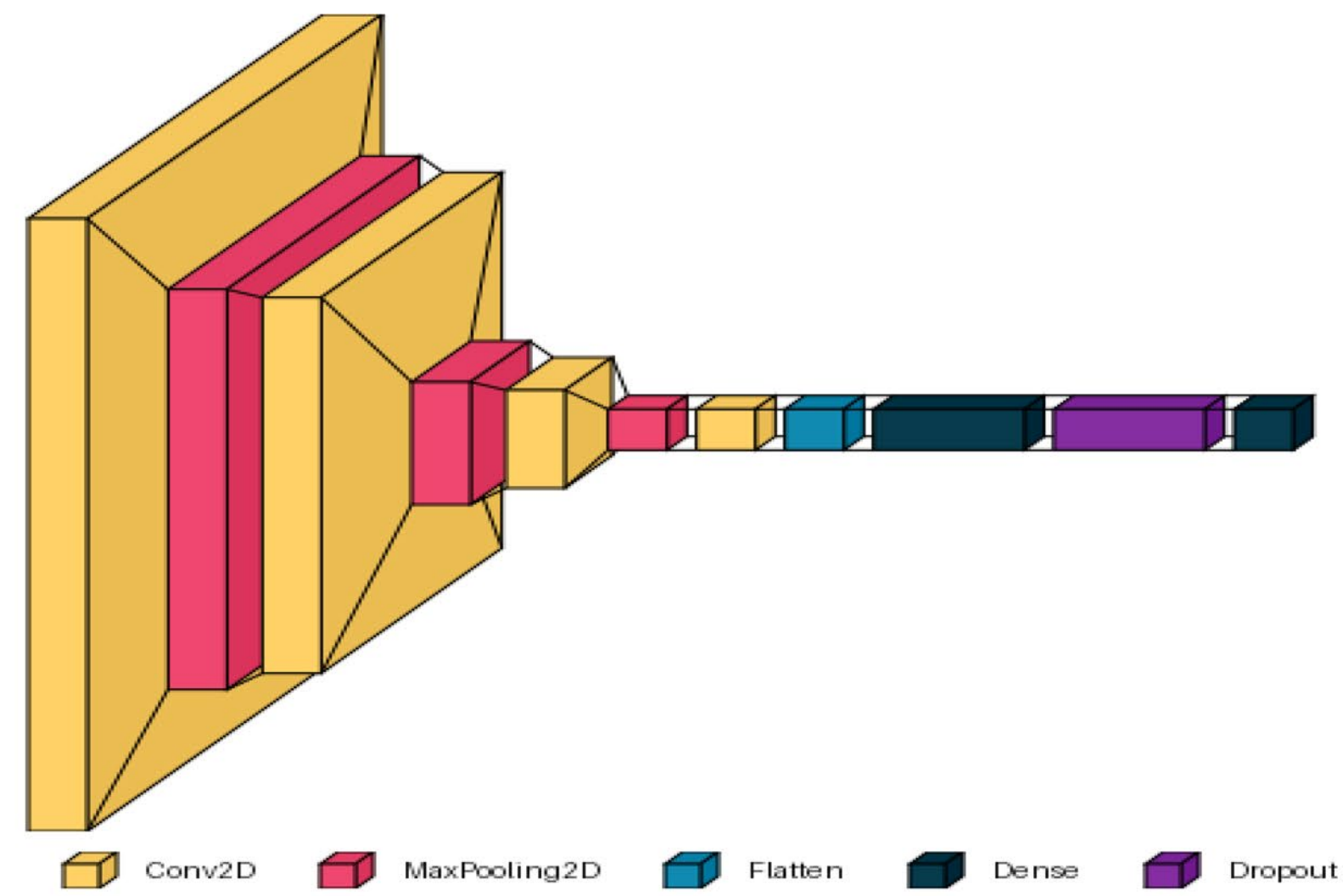

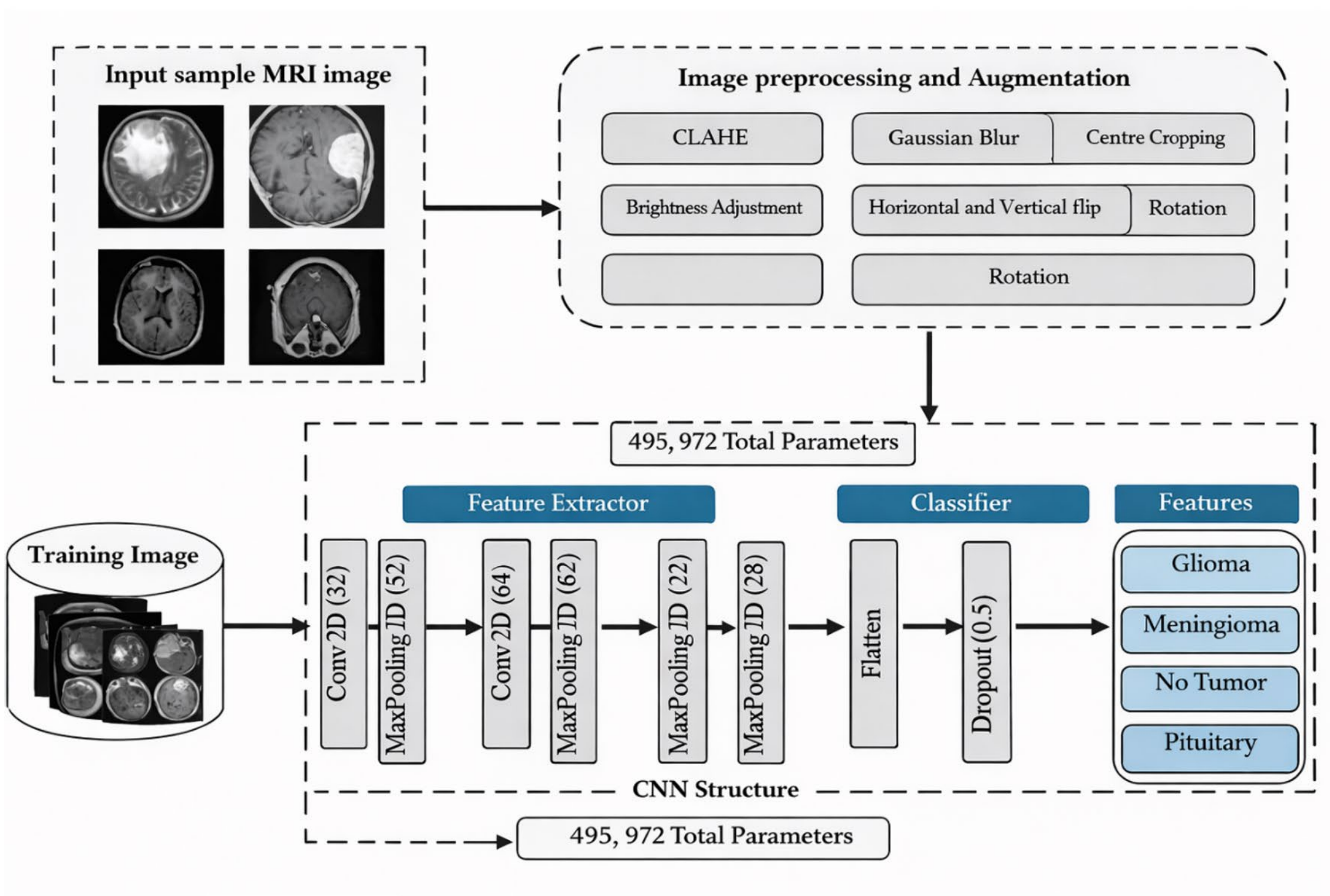

**Fig. 5** **A** visualization of the CNN for feature extraction. **B** CNN feature extraction layer

$$E = Conv2D(\boldsymbol{I}, kernel_size = PATCH_SIZE, strides = PATCH_SIZE) \quad (2)$$

The result is reshaped into a sequence of embeddings $X \in R^{N \times d}$, where $N$ = the number of patches and where $d$ = the dimensional feature vector (here, d = 32). To preserve the spatial arrangement of these patches, learnable positional embeddings are added to the patch representations. The embedded patch sequence is subsequently processed through a stack of four transformer blocks. Each block consists of a multi-head self-attention module with 4 attention heads. This mechanism allows the model to relate information from different patches. The attention operation is formulated as Eq. 3:

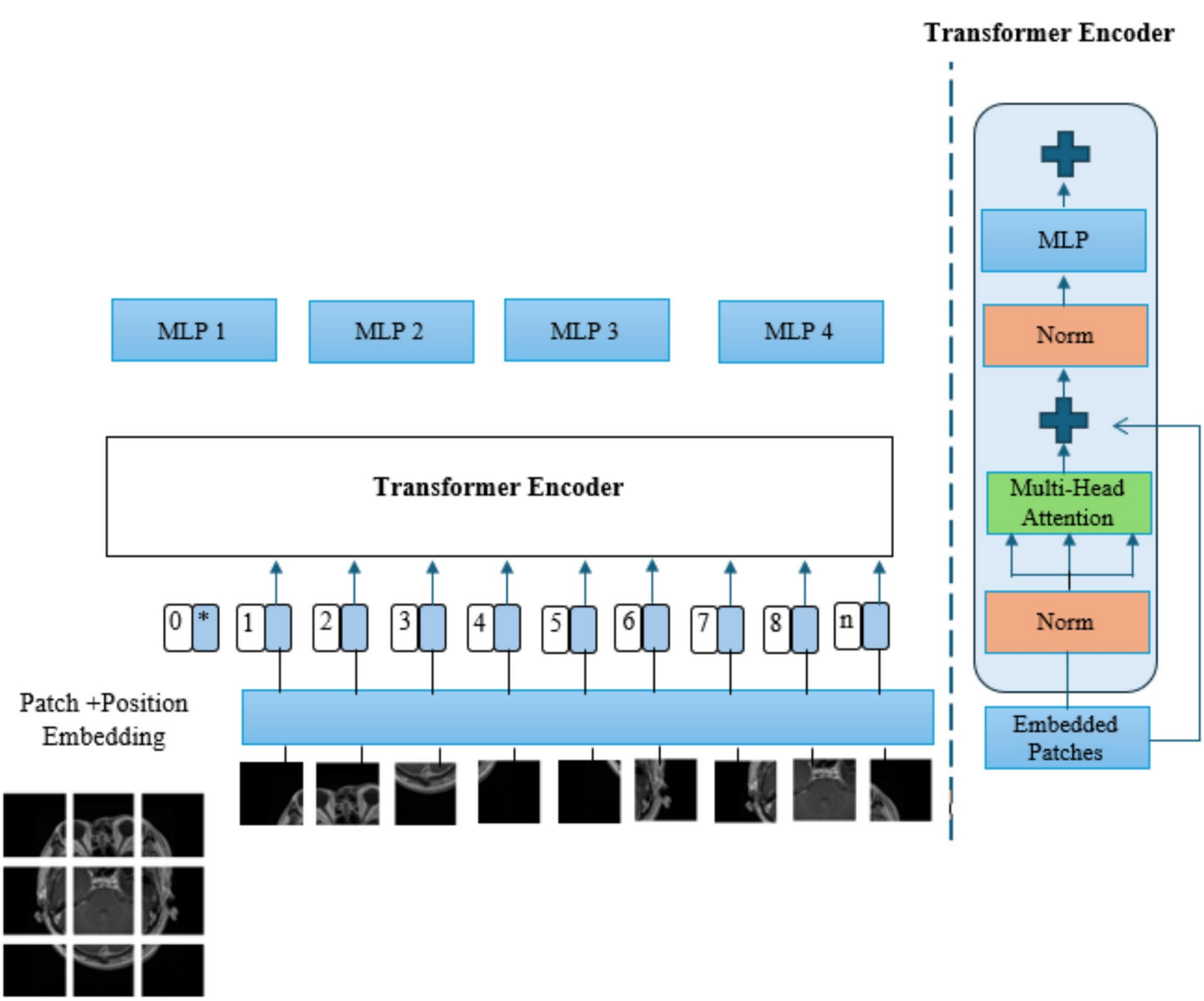

**Fig. 6** ViT architecture for feature extraction

$$Attention(Q, K, V) = softmax(\frac{QK}{\sqrt{d_k}})V \quad (3)$$

where queries (Q), keys (K), and values (V) are derived from the input, and $d_k$ Is the dimension per attention head.

After the attention layer, a feed-forward network (FFN) expands the embedding dimension by a factor of four before projecting it back to its original size. The attention and FFN sublayers are interleaved with layer normalization and utilize residual connections, ensuring stable gradient flow and practical learning. After the transformer layers, the output is normalized, flattened, and passed through a dense layer with 512 neurons (ReLU activation) to align the dimensionality with that produced by the CNN branch:

$$f_{ViT} = ReLU(W_{vit} \cdot Flatten(LayerNorm(X)) + b_{vit}). \quad (4)$$

This global representation helps capture relationships across the entire image, providing context that CNNs alone might miss.

#### *4.1.3 GRU layer*

The final stage of the GRU layer classifies the features (see Fig. 7). The GRU is employed to model pseudo-temporal dependencies derived from the spatial arrangement of features extracted by the CNN and ViT. While the input MRI slices are static, the features extracted across multiple regions form a meaningful spatial progression. The GRU, though typically used for temporal sequences, is theoretically appropriate here because it captures patterns in the ordered feature vector. This ordering can represent spatial relationships, such as how features progress from one region of the brain to another. Unlike deeper feed-forward networks, the GRU explicitly models dependencies—helping to capture subtle spatial patterns that evolve across the extracted feature sequence.

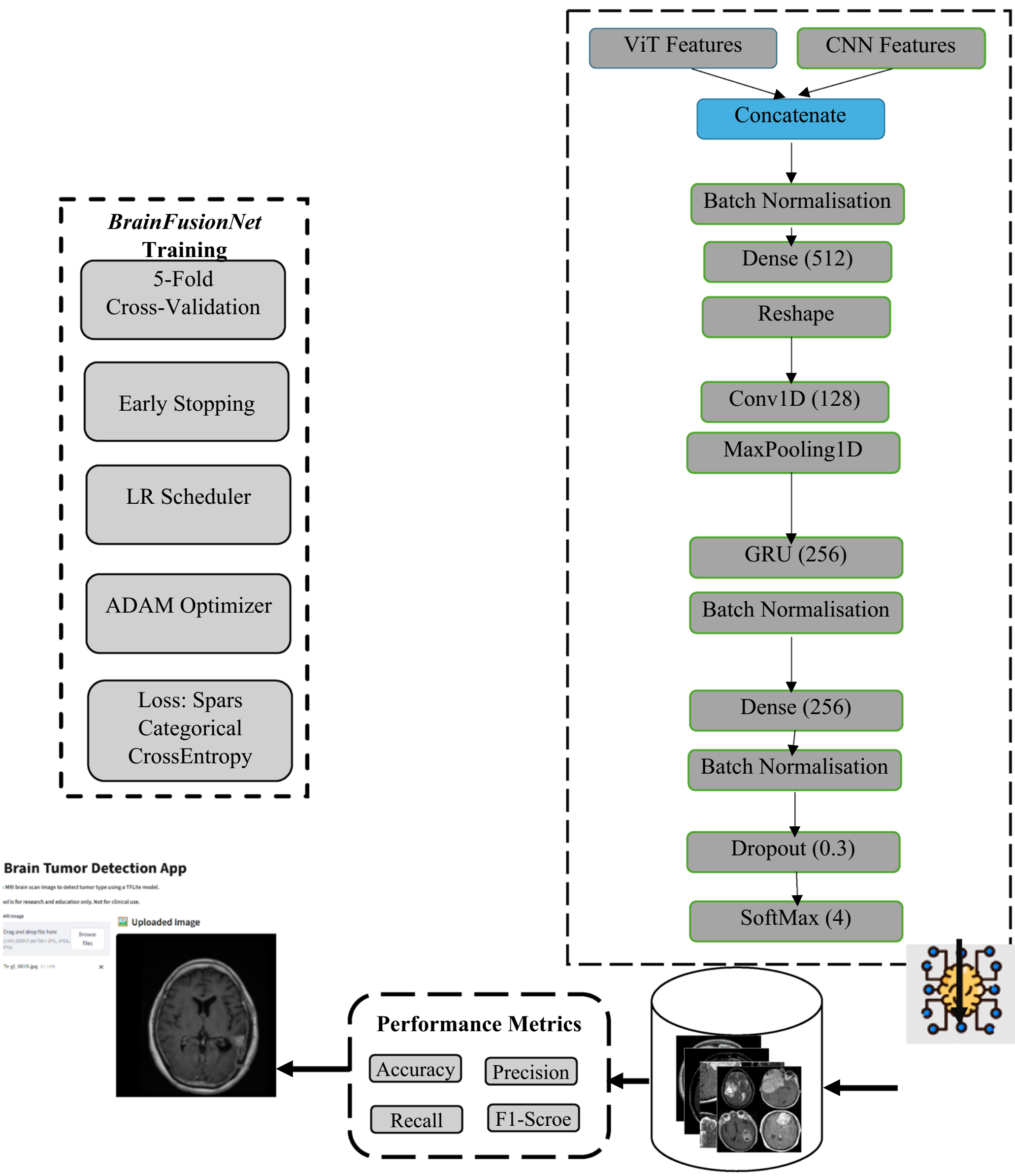

**Fig. 7** GRU for MRI classification

The CNN and ViT layers produce a 512-dimensional vector, and the GRU concatenates to form a unified feature vector:

$$f_{combined} = [f_{CNN}, f_{ViT}] \in R^{1024} \tag{5}$$

The feature vector is reshaped into a sequential format to facilitate the modeling of temporal dependencies. A one-dimensional convolutional layer (Conv1D) with 128 filters and a kernel size of 3 is applied to capture local temporal patterns within the sequence, followed by a MaxPooling1D layer to reduce dimensionality. The processed sequence is then fed into a gated recurrent unit (GRU) layer with 256 units and a recurrent dropout rate of 0.2. This recurrent processing allows the model to capture sequential dependencies corresponding to subtle temporal or spatial patterns in the MRI data.

The GRU output is further refined through batch normalization and dropout (rate of 0.3) before being fed into a fully connected layer with 256 neurons (ReLU activated). Finally, a Softmax output layer produces the class probabilities over the four tumor categories. The overall mapping of the integrated model can be expressed as:

$$y = softmax(f_{FC}(GRU(Conv1D(Reshape(Dense(BN(f_{combined})))))) \tag{6}$$

where $BN$ represents batch normalization, and where $f_{FC}$ denotes the series of fully connected layers after GRU processing.

### 4.2 Training *BrainFusionNet*

As suggested by Apicella (2025), this study applied SMOTE (Synthetic Minority Over-sampling Technique) and Data Augmentation after splitting the dataset into training, validation, and testing sets. This ensures that no information from the test set was used during the training phase, which is a critical step to prevent data leakage. SMOTE was applied only to the training set in each fold to balance class distribution. This was done by generating synthetic samples from the existing training data, without affecting the test set. Data augmentation techniques (e.g., random rotations, flips, and brightness adjustments) were applied only to the training data. These transformations are designed to increase the diversity of the training set without introducing any information from the test set.

The *BrainFusionNet* model was trained using K-fold stratified cross-validation. To ensure that each fold trains the same proportion of class labels as in the original dataset, K-5 stratified cross-validation was selected. In this study, 70% of the images were used for training *BrainFusionNet*, 20% for validation, and 10% for testing its performance on unseen test images. Since the study utilised two (2) brain tumor datasets, two separate training sets were created for each dataset. Hence, test results also emerged from two distinct datasets.

To prevent data leakage, the images were transferred to the train, validation, and test directories. Separating the images into three directories ensured against any potential data leakage, and hence, the test and validation sets were not considered the same. As Ashmore et al. (2021) suggested, the *BrainFusionNet* model verification stage included evaluating the final model on a test dataset (10% of the images) that was not used during training.

As suggested by Apicella (2025), this study applied SMOTE (Synthetic Minority Over-sampling Technique) and Data Augmentation after splitting the dataset into training and testing sets. This ensures that no information from the test set was used during the training phase, which is a critical step to prevent data leakage. SMOTE was applied only to the training set in each fold to balance class distribution. This was done by generating synthetic samples from the existing training data, without affecting the test set. Data augmentation techniques (e.g., random rotations, flips, and brightness adjustments) were applied only to the training data. These transformations are designed to increase the diversity of the training set without introducing any information from the test set. The training hyperparameters are provided in Table 2.

The *Adam* algorithm was used for model optimization. The algorithm ensures the model weights adapt to the calculated gradients, thereby enhancing performance and accuracy in classifying brain tumor modalities. The loss function was sparse categorical cross-entropy. The parameters of CNN, ViT, and GRU are provided in Table 3, and training hyperparameters are provided in Table 2.

**Table 2** Parameters of *BrainFusionNet*

| Parameters | Trainable | Nontrainable |
|---|---|---|
| CNN | 493,920 | 0 |
| Vit | 1,762,144 | 0 |
| GRU | 3,347,012 | 3072 |

### 4.3 *BrainFusionNet* performance

The performance of the *BrainFusionNet's* models in brain tumor detection and classification is evaluated using the following metrics:

$$Accuracy = ((TP + TN)) / ((TP + FP + FN + TN))$$

$$Precision = TP/(TP + FP)$$

$$Recall = TP/(TP + FN)$$

$$F1score = 2 \times (Precision \times Recall) / (Precision + Recall)$$

$$Specificity = TN/(TN + FP)$$

Furthermore, the accuracy curve, data loss curve, and confusion matrix were used to visualize the models' performance.

Table 4 shows an average accuracy of 98% for both the 3-class and 4-class tasks using *BrainFusionNet*. Similar accuracy indicates that the features learned by the *BrainFusionNet* model are sufficient to detect and classify brain tumor modalities from MRI images.

Table 5 presents detailed classification metrics, including precision, recall, and F1 score of 3-class and 4-class MRI. In the 3-class setting, Glioma and Meningioma achieve higher accuracy than the Pituitary tumors class (precision, recall, and F1 score = 1.0). The macro- and weighted averages also confirm balanced performance across all classes, with an overall accuracy of 98%, reflecting the strong results observed in the confusion matrix.

In the 4-class MRI image, *BrainFusionNet* achieves higher classification accuracy in the No_tumor and Pituitary classes. However, Meningioma has a slightly lower Recall (0.96). However, the model maintains 98% accuracy, comparable to that of the 3-class MRI classification.

The confusion matrix (Fig. 8) shows that in the 3 classes, the model correctly predicted glioma, meningioma, and pituitary, achieving high correct prediction counts of 1246, 1121, and 1182, respectively. Even in the 4-class confusion matrix, Pituitary and No_tumor

**Table 3** Hyperparameters of training

```
Epochs = 50
Batch size = 16
Image size = (64, 64, 3)
Learning rate = 1.0000e-04
K_folds = 5
Optimizer = Adam(learning_rate=LEARNING_RATE)
Loss = SparseCategoricalCrossentropy(from_logits=True)
Early stopping = EarlyStopping(monitor='val_accuracy', patience=10, verbose=1,
restore_best_weights=True)
Learning_rate_scheduler = LearningRateScheduler(lambda epoch: LEARNING_RATE * 0.1 **
(epoch//10))
Callbacks = [early_stopping, lr_scheduler]
```

1: Initialize *BrainFusionNet* with three branches:

- CNN branch: 6 convolutional layers (32, 64, 128 filters) with ReLU and max pooling, output 512-dimensional vector
- ViT branch: Patch embedding, 4 transformer blocks, output 512-dimensional vector
- GRU branch: Sequential modeling for classification (described below)

```
2: For fold k = 1 to 5 do
3:     Split dataset D into train, validation, and test using k-fold cross-validation
4:     Reset model weights
5:     Compile BrainFusionNet with Adam optimizer and appropriate loss function
6:     For each epoch = 1 to N_epochs do
7:         For each batch (x, y) in train data do
8:             CNN features = CNN branch(x)
9:             ViT features = ViT branch(x)
10:            Combine features = concatenate(CNN features, ViT features)
11:            Reshape combined features into a sequence
12:            Sequence = Conv1D(Sequence)
13:            Sequence = MaxPooling1D(Sequence)
14:            GRU output = GRU(Sequence) with recurrent dropout
15:            GRU output = batch_normalization(GRU output)
16:            GRU output = dropout(GRU output)
17:            Dense output = Dense layer with ReLU(GRU output)
18:            Predictions = Softmax output(Dense output)
19:            Calculate loss and apply gradients
20:        End for batch
21:    End for epoch
22: Evaluate model on validation data
23: Save validation accuracy and generate XAI outputs:
    - LIME for local feature importance
    - SHAP for feature contributions
    - Grad-CAM heatmaps for visual relevance
24: End for fold
25: Compute average validation accuracy across all folds
26: Evaluate on unseen test data
```

**Table 4** *BrainFusionNet* combined result

| 3-Class (%) | 4-Class |
|---|---|
| Accuracy: 98.15 | Accuracy: 98.16% |
| Precision: 98.15 | Precision: 98.16% |
| Recall: 98.15 | Recall: 98.16% |
| F1-Score: 98.15 | F1-Score: 98.16% |

**Table 5** Combined classification report

| Class | Precision | Recall | F1-Score | Support |
|---|---|---|---|---|
| Glioma | 0.97 | 0.98 | 0.98 | 1274 |
| Meningioma | 0.97 | 0.97 | 0.97 | 1156 |
| Pituitary | 1 | 1 | 1 | 1186 |
| Accuracy | – | – | 0.98 | 3616 |
| Macro Avg | 0.98 | 0.98 | 0.98 | 3616 |
| Weighted Avg | 0.98 | 0.98 | 0.98 | 3616 |
| **Class** | **Precision** | **Recall** | **F1-Score** | **Support** |
| Glioma | 0.98 | 0.98 | 0.98 | 1758 |
| Meningioma | 0.97 | 0.96 | 0.96 | 1748 |
| No_tumor | 0.99 | 0.99 | 0.99 | 1808 |
| Pituitary | 0.98 | 1 | 0.99 | 1806 |
| Accuracy | | | 0.98 | 7120 |
| Macro Avg | 0.98 | 0.98 | 0.98 | 7120 |
| Weighted Avg | 0.98 | 0.98 | 0.98 | 7120 |

are classified with the highest accuracy, with counts of 1800–1792, respectively. Minor misclassifications occur between Glioma and Meningioma. Especially some meningioma images misclassified as No_tumor or Pituitary, which might be due to overlapping visual characteristics. Figure 9 presents the fold-wise accuracy and data-loss curves.

The confusion matrices (Figs. 10A-B) for the five folds of *BrainFusionNet* on 3-class and 4-class brain tumors show consistency with minimal variance across folds.

The ROC curves (Figs. 11. A-B) for each class across the fivefold validation reveal that *BrainFusionNet* demonstrates the area under the curve (AUC) values close to or equal to 1.00 for all classes. The curves indicate that the model can distinguish between the tumors. The steep rise near the y-axis and the flat curves near the top confirm low false-positive rates and high true-positive rates. Overall, the consistently high AUCs across all the folds reflect the strong generalizability and robustness of *BrainFusionNet* in multiclass brain tumor classification.

### 4.4 Additional ablation studies for CNN

To examine the stability and performance of the M-Net model, we conducted two ablation studies on the 4-class brain tumor. Since this dataset achieved the highest Accuracy, precision, and Recall, we selected it for the ablation study. In this research, three (3) configurations of the M-Net architecture were eliminated: the first Convolutional Layer, the last Convolutional Layer, and the Dropout (0.5) layer, to test whether regularization helps in this case. The aim was to determine whether eliminating components of M-Net affects performance and to demonstrate that the M-Net design is efficient and adaptable for addressing the challenges of brain tumor classification. The results (see Table 6) reveal that changing the architecture decreases the performance of the original M-Net model by 1%.

#### *4.4.1 Ablation configuration: removing pooling layers*

Pooling layers are generally used to reduce dimensionality while retaining important spatial information. Removing them will help you test how critical they are for the CNN model Table 7.

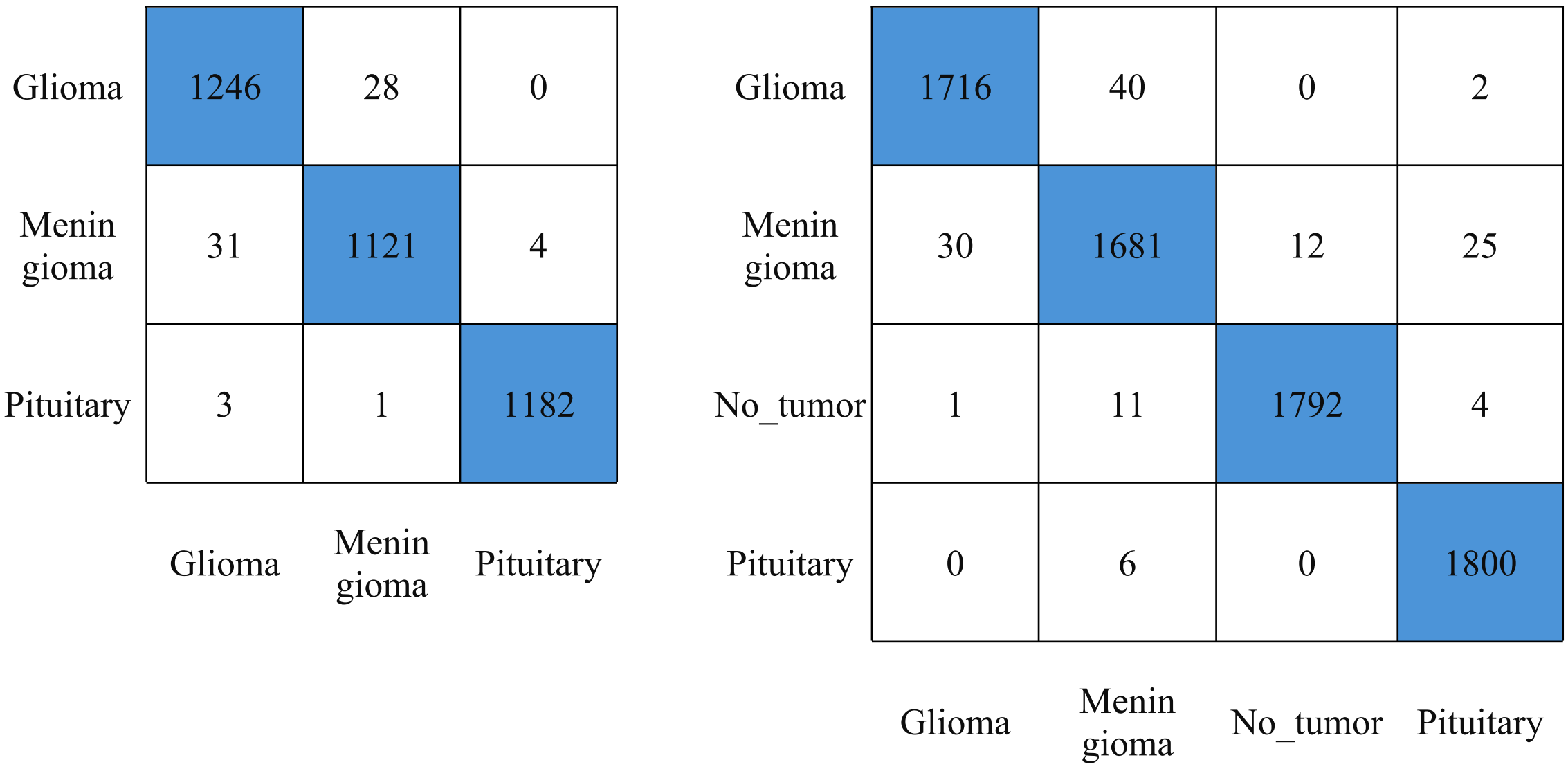

**Fig. 8** Combined confusion matrix

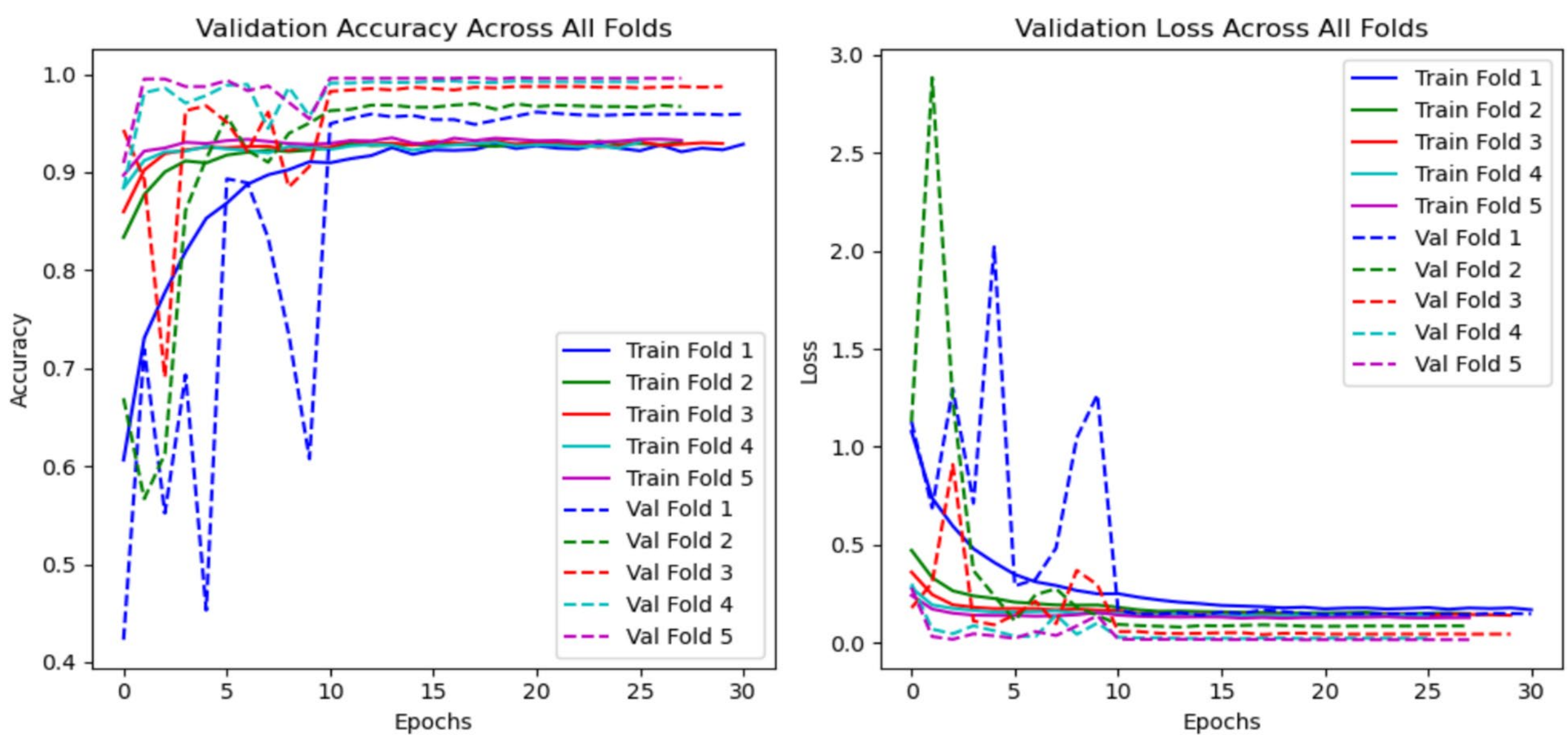

**Fig. 9** Fold-wise Accuracy and loss curves, Dataset A (fivefold cross-validation)

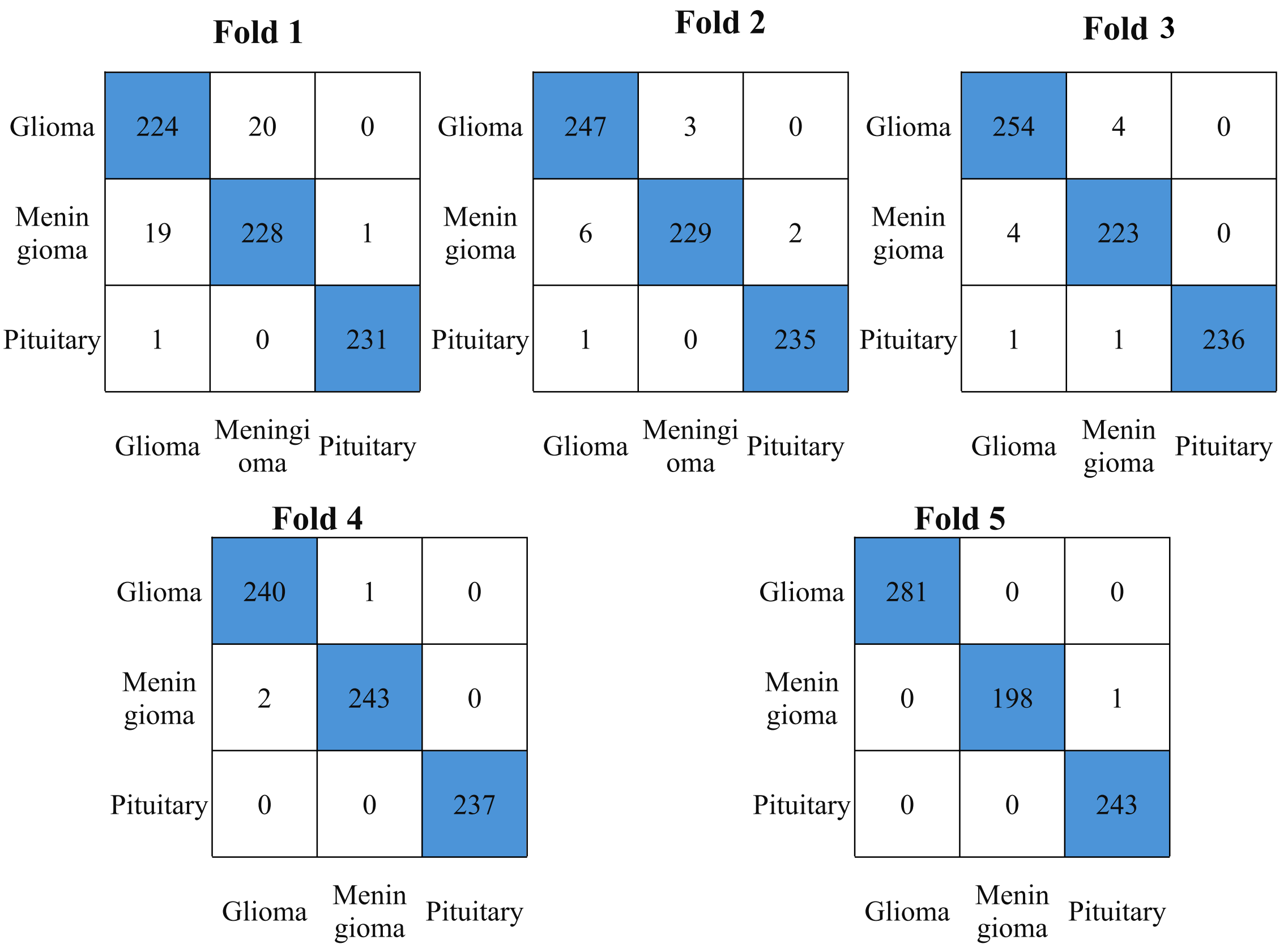

**Fig. 10** **A** Confusion matrix of Dataset A (fivefold cross-validation. **B** Confusion matrix of Dataset B (fivefold cross-validation) using *BrainFusionNet*

#### *4.4.2 Ablation configuration: effect of different kernel sizes*

This experiment tests how varying the kernel size affects model performance. Larger kernels capture more global information, while smaller kernels focus on finer details Table 8.

#### *4.4.3 Ablation configuration:varying the kernel sizes in the convolutional layers (e.g., from 3 × 3 to 5 × 5 and 7 × 7)*

#### *4.4.4 Ablation configuration: effect of removing dropout regularization and dropout layer*

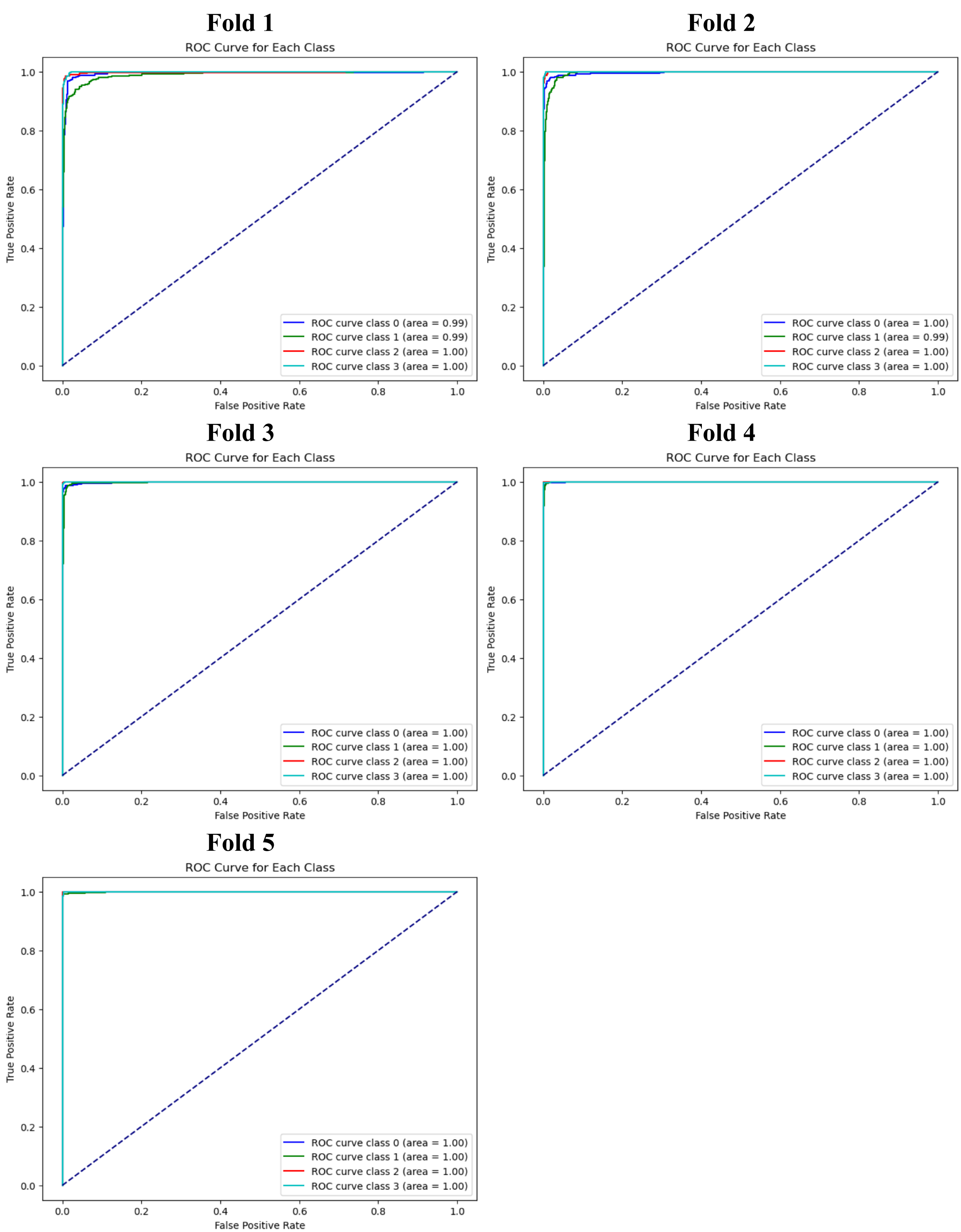

**Fig. 11** **A**: ROC curve of Dataset A (fivefold cross-validation). **B**: ROC curve of Dataset B (fivefold cross-validation)

**Table 6** Ablation study results for pooling layer

| Ablation strategy | Training accuracy (%) | Validation accuracy (%) | Test accuracy (%) |
|---|---|---|---|
| Original (with Pooling) | 99 | 98 | 97 |
| Without pooling | 98 | 97 | 95 |

**Table 7** Ablation study results for kernel sizes

| Ablation strategy | Training accuracy (%) | Validation accuracy (%) | Test accuracy (%) |
|---|---|---|---|
| Original (3×3 kernels) | 99 | 98 | 97 |
| 5×5 kernels | 99 | 97 | 96 |
| 7×7 kernels | 98 | 96 | 94 |

**Table 8** Ablation study results for dropout layer

| Ablation strategy | Training accuracy (%) | Validation accuracy (%) | Test accuracy (%) |
|---|---|---|---|
| Original (with Dropout) | 99 | 98 | 97 |
| Without dropout | 100 | 97 | 95 |

**Table 9** Ablation study results for patch sizes

| Patch size | Training accuracy (%) | Validation accuracy (%) | Test accuracy (%) |
|---|---|---|---|
| Original (9 patches) | 99 | 97 | 96 |
| 16 patches | 98 | 96 | 95 |
| 64 patches | 98 | 97 | 97 |

**Table 10** Ablation study results for transformer depths

| Transformer depth | Training accuracy (%) | Validation accuracy (%) | Test accuracy (%) |
|---|---|---|---|
| Original (12 layers) | 99 | 97 | 96 |
| 6 layers | 98 | 96 | 95 |
| 18 layers | 100 | 97 | 96 |

This study investigates the impact of removing Dropout layers on the model's performance, especially in terms of overfitting and generalization.

The results show that removing the Dropout layer improves training accuracy. However, the model's performance on the validation and test sets drops, indicating that Dropout plays a significant role in regularization and reducing overfitting Table 9.

#### *4.4.5 Additional ablation studies for ViT ablation configuration: effect of changing patch size*

This experiment examines the impact of varying patch size on the performance of the Vision Transformer (ViT). Patch size influences the receptive field and determines how the model captures spatial information from the image Table 10.

#### *4.4.6 Ablation studies using different transformer depths*

This study investigates how varying the number of transformer layers affects the performance of the Vision Transformer (ViT). The transformer's depth controls how much contextual information it captures from the input data Table 11.

**Table 11** Ablation study results for attention heads

| Attention heads | Training accuracy (%) | Validation accuracy (%) | Test accuracy (%) |
|---|---|---|---|
| Original (12 heads) | 99 | 97 | 96 |
| 8 heads | 98 | 96 | 94 |
| 16 heads | 99 | 98 | 97 |

**Table 12** Ablation study results for CNN and ViT architectures

| Architecture | Ablation strategy | Training accuracy (%) | Validation accuracy (%) | Test accuracy (%) |
|---|---|---|---|---|
| CNN | Remove pooling | 98 | 97 | 95 |
| CNN | Vary kernel size | 99 | 97 | 96 |
| CNN | Remove dropout | 100 | 97 | 95 |
| ViT | Change patch size | 98 | 96 | 95 |
| ViT | Vary transformer depth | 98 | 96 | 95 |
| ViT | Change attention heads | 99 | 98 | 97 |

#### *4.4.7 Ablation configuration: test with 8, 12, and 16 attention heads.*

This study examines how varying the number of attention heads affects the performance of the Vision Transformer (ViT). The number of attention heads determines how the self-attention mechanism distributes attention across different parts of the input Table 12.

The ablation studies clearly demonstrate the significance of each architectural component in both the CNN and ViT models. The results suggest removing pooling or dropout layers reduces performance, highlighting the importance of regularization and spatial reduction in feature extraction. Adjusting patch size, transformer depth, and the number of attention heads all influence performance. Notably, smaller patch sizes and fewer attention heads often lead to a loss of model performance, indicating that a balanced configuration is essential for optimal feature extraction and attention mechanisms.

These results validate *BrainFusionNet's* efficacy and the thoughtful design choices in its construction, aligning with claims about the effectiveness of feature fusion. This analysis could be added to the methods or results section of your paper for clarity and rigor.

### 4.5 Benchmark comparison

This study applies six (6) SOTA CNNs to the 4-class brain tumor MRI dataset. The selection of CNNs was based on the taxonomy of Khan et al. [38]. The selection aimed to cover most CNN architectures to investigate the most effective MRI image detection and classification network. The six SOTA CNNs were selected from multipath (DenseNet121, ResNet50), depthwise (InceptionV3), width-based multiconnection (Xception), depthwise separable (MobileNet), and spatial-exploitation (VGG16) networks. Three optimizers, Adam, Adamax, and RMSprop, were also used to find an effective optimizer in CNN.

**Table 13** Performance comparison of the SOTA CNNs and optimizers

| Model | Epoch (Adam) | Accuracy (Adam) | Epoch (Adamax) | Accuracy (Adamax) | Epoch (RMSprop) | Accuracy (RMSprop) |
|---|---|---|---|---|---|---|
| DenseNet121 | 11 | 0.98 | 37 | 0.96 | 23 | 0.98 |
| ResNet50 | 37 | 0.97 | 46 | 0.89 | 43 | 0.97 |
| InceptionV3 | 26 | 0.97 | 36 | 0.88 | 37 | 0.97 |
| Xception | 27 | 0.97 | 30 | 0.94 | 33 | 0.95 |
| MobileNet | 35 | 0.89 | 34 | 0.73 | 27 | 0.87 |
| VGG16 | 30 | 0.92 | 35 | 0.84 | 36 | 0.96 |

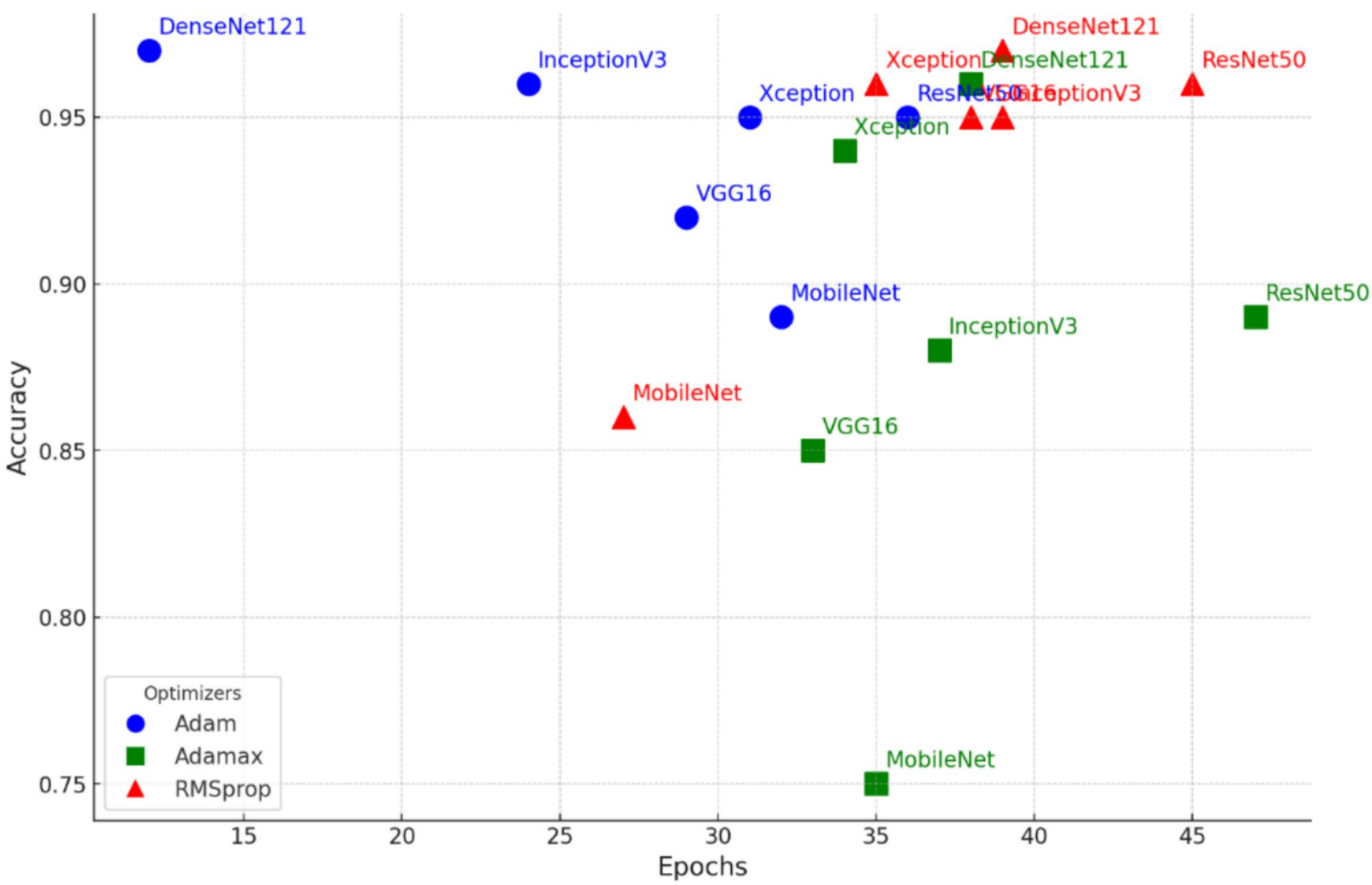

**Fig. 12** SOTA CNN models, accuracy, and epochs of the 4-class dataset

Table 13 compares the performance of the six SOTA CNNs when using the Adam, Adamax, and RMSprop optimizers. DenseNet121 achieves the highest accuracy of 98% when both Adam and RMSprop are used. However, Adam requires 11 and an RMSprop of 23 epochs. Conversely, ResNet50 requires significantly more epochs; however, optimizers do not impact the accuracy. MobileNet and VGG16 exhibit lower accuracy, particularly with Adamax (73–84%, respectively). Most importantly, Adam and RMSprop show better performance consistency, whereas Adamax appears less efficient in terms of convergence and accuracy, particularly on MobileNet and InceptionV3. A scatter plot is also presented in Fig. 12 to illustrate the relationship between accuracy and the number of epochs for the Adam, Adamax, and RMSprop optimizers across the six SOTA CNNs.

Figure 13 also presents a scatter plot illustrating the relationship between accuracy and epochs for the Adam, Adamax, and RMSprop optimizers on six SOTA CNNs.

The confusion matrix (Fig. 14) suggests that ResNet50 and DenseNet121 exhibit lower Type 1 and Type 2 errors. The lower Type I and Type II values indicate better overall classification performance of DenseNet121 and ResNet50. InceptionV3 also shows competitive performance, while Xception balances false positives and false negatives. However, MobileNetV2 and VGG16 have more false positives and false negatives. Overall, DenseNet121 and ResNet50 appear to be the most robust

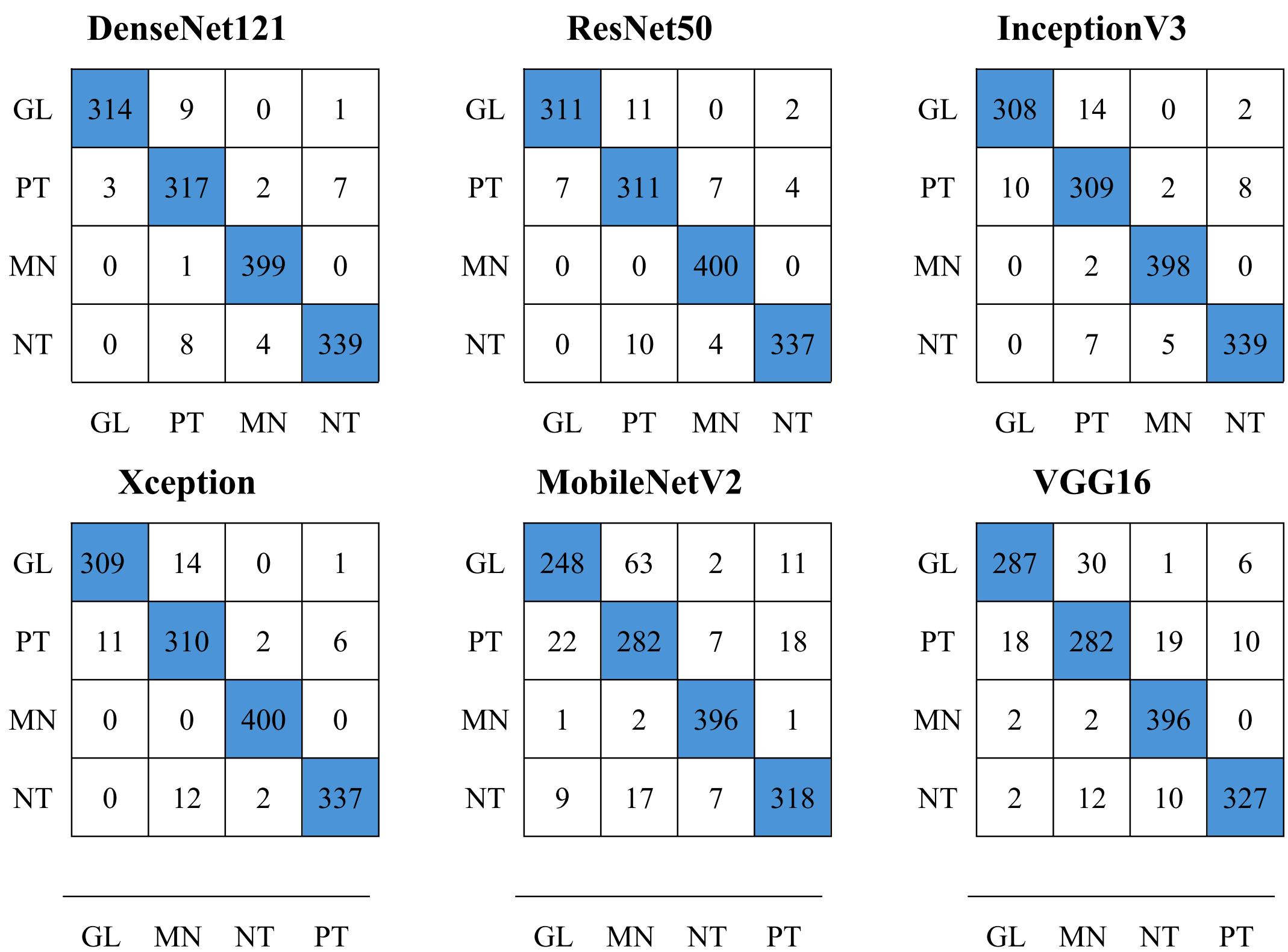

**Fig. 13** Confusion matrix of the SOTA CNNs for the 4-class dataset

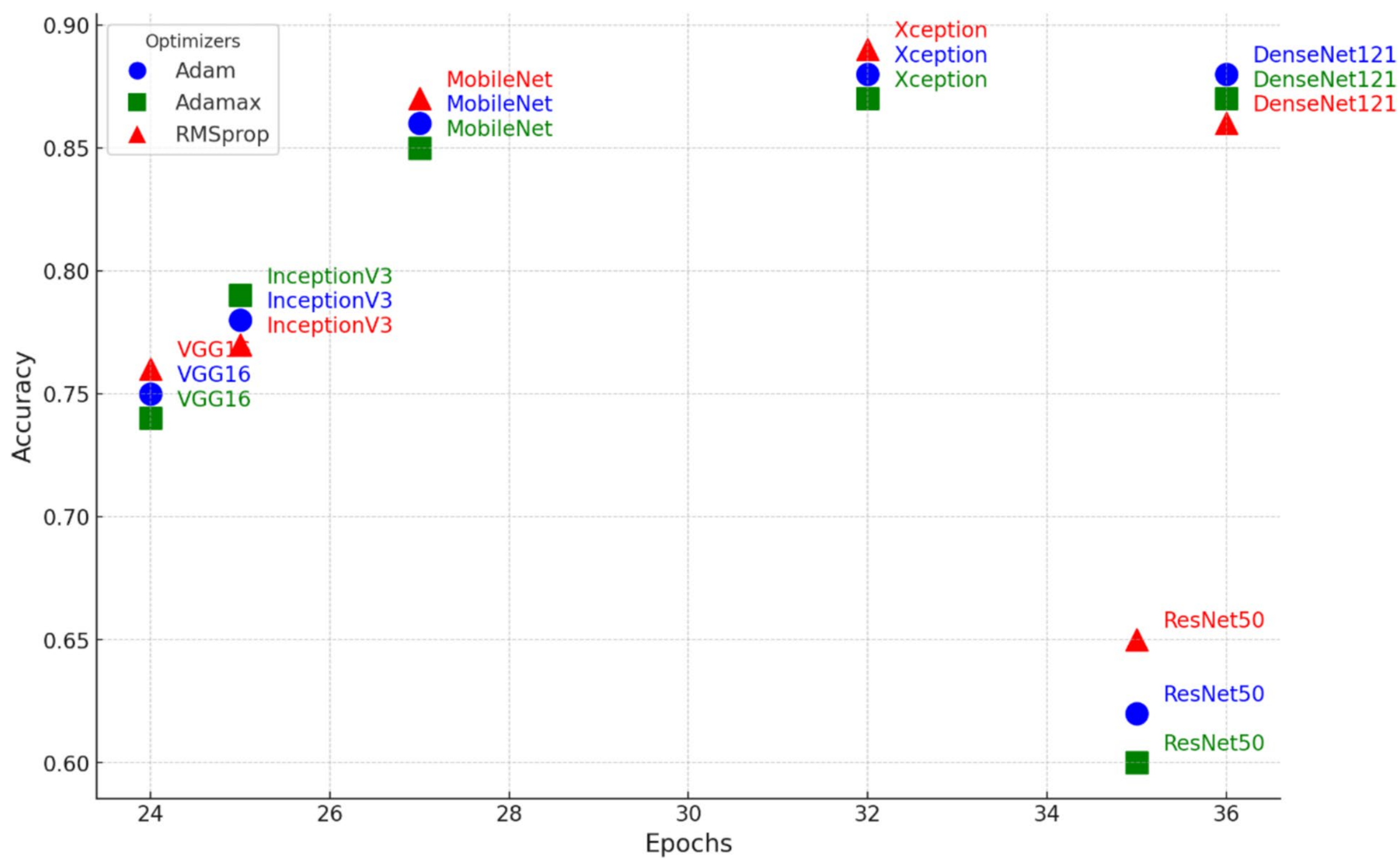

**Fig. 14** Transfer learning CNN models, accuracy, and number of epochs

**Table 14** Performance comparison of the SOTA CNNs' transfer learning and optimizers

| | Epoch | Accuracy (Adam) | Epoch | Accuracy (Adamax) | Epoch | Accuracy (RMSprop) |
|---|---|---|---|---|---|---|
| DenseNet121 | 36 | 0.84 | 36 | 0.84 | 36 | 0.84 |
| ResNet50 | 35 | 0.59 | 35 | 0.59 | 35 | 0.59 |
| InceptionV3 | 25 | 0.8 | 25 | 0.8 | 25 | 0.8 |
| Xception | 33 | 0.87 | 33 | 0.87 | 33 | 0.87 |
| MobileNet | 25 | 0.87 | 25 | 0.87 | 25 | 0.87 |
| VGG16 | 23 | 0.75 | 23 | 0.75 | 23 | 0.75 |

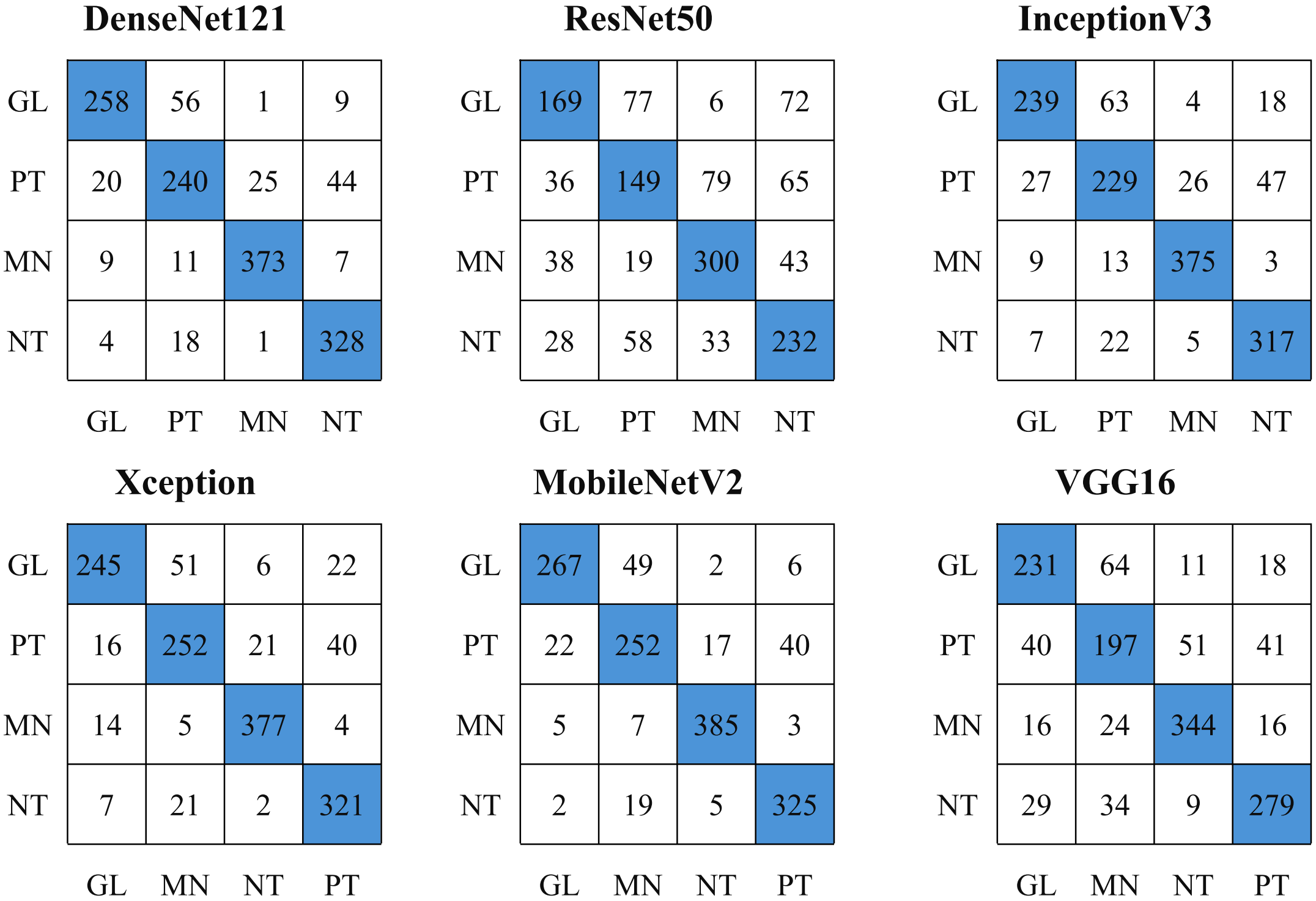

**Fig. 15** Confusion matrix of transfer learning

models, while MobileNetV2 shows weaker performance, with higher misclassification rates.

#### 4.6 Transfer learning performance

The accuracy table and scatter plot suggest that DenseNet121 achieves 84% accuracy across all optimizers (Table 14 and Fig. 12). Xception closely follows, reaching 87% accuracy with 32 epochs. ResNet50 achieves the lowest accuracy in transfer learning (59%), despite requiring ~ 35 epochs, suggesting optimization challenges.

The confusion matrix (see Fig. 15) for transfer learning with DenseNet121 and MobileNetV2 shows lower Type 1 and Type 2 errors, suggesting stronger classification performance. InceptionV3 and Xception maintain balanced performance, with moderate false positives and negatives. VGG16, however, struggles with higher misclassification rates, especially in distinguishing between classes, making it less reliable than the other methods. ResNet50 has the highest misclassification rates, with significant false positives and false negatives.

#### 4.7 Results of XAI

This study uses three (3) explainable AI techniques, namely, LIME, SHAP, and Grad-CAM, to generate local and global explanations for the *BrainFusionNet* predictions for validation. The following sections present discussions related to XAI.

##### *4.7.1 Visualization via LIME*

LIME-generated heatmaps showing which regions (superpixels) positively or negatively contributed to the predicted class (see Fig. 16). The red and green marks generated via LIME explain the input image and thus assist in understanding the process of determining the brain tumor modalities.

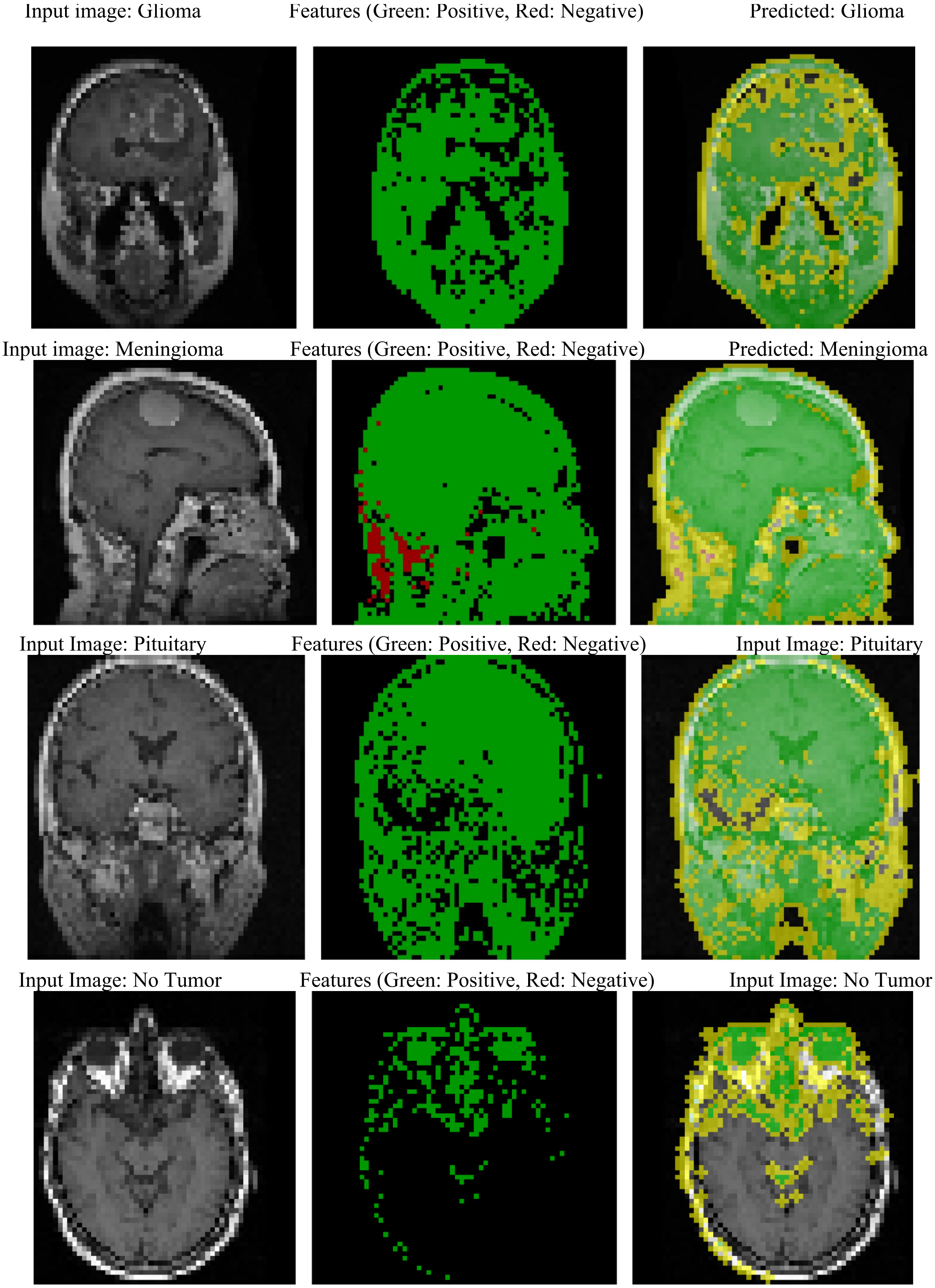

**Fig. 16** LIME partition explainer of MRI images

#### *4.7.2 Visualization via SHAP*

SHAP explanations are generated via a pretrained BrainTumor.h5 (see Fig. 17). The SHAP explainer (shap.Explainer) is initialized with the model and masker to compute Shapley values. This SHAP visualization explains the model's predictions in a brain tumor classification task by highlighting the contributions of specific MRI regions to each class. The red regions indicate features that positively influence the prediction of a class, whereas the blue regions indicate features that reduce the likelihood.

#### *4.7.3 Grad-CAM analysis of correctly/incorrectly classified tumor modalities*

The Grad-CAM generates heatmaps to visualize which regions of an image contributed most to the *BrainFusionNet's* prediction. Figure 18 shows the Grad-CAM visualizations of the model's incorrect predictions across various brain tumor cases. The red on the maps denotes greater attention given to those locations, whereas the blue denotes less attention to those regions. The heatmaps reveal that the model often focuses on irrelevant or nontumor areas, leading to misclassifications, such as "No_tumor" instead of "Pituitary" or "Glioma"; however, in the correctly classified image. Figure 19 shows the Grad-CAM's focus on relevant features for its correct prediction.

Grad-CAM was also used to generate pixel intensity (see Figs. 20 and 21). The right panel displays the gradient × input explanation, where the pixel intensities indicate how much each part of the image contributed to the model's decision. Bright areas have a greater influence on the prediction. In this case, the highlighted regions in the Gradient × Input explanation do not align well with the actual tumor, indicating that the model focused on irrelevant areas and missed critical tumor features. This suggests that a misalignment between the model's learned features and the critical tumor characteristics is necessary for accurate classification.

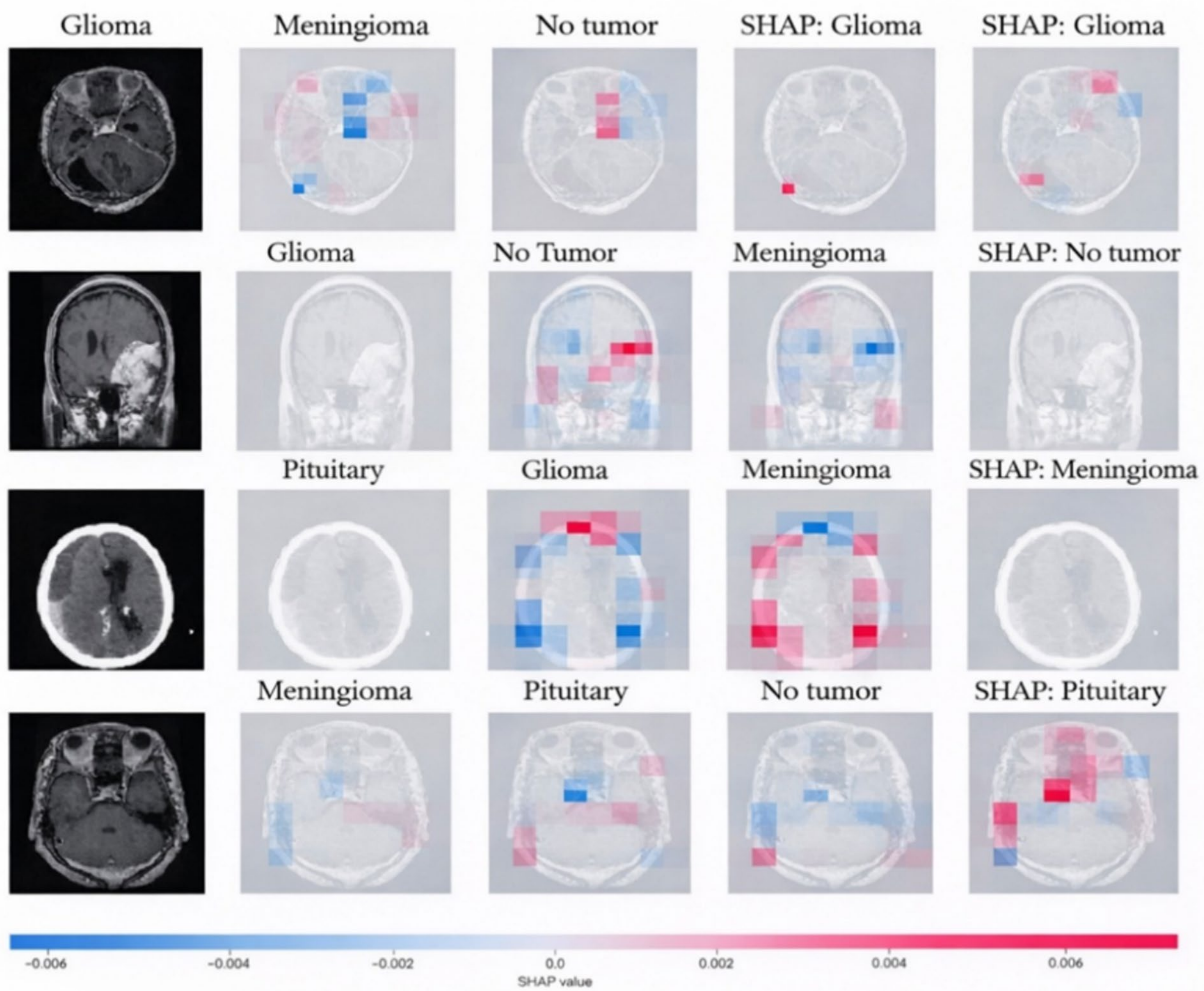

**Fig. 17** SHAP explainer of MRI images

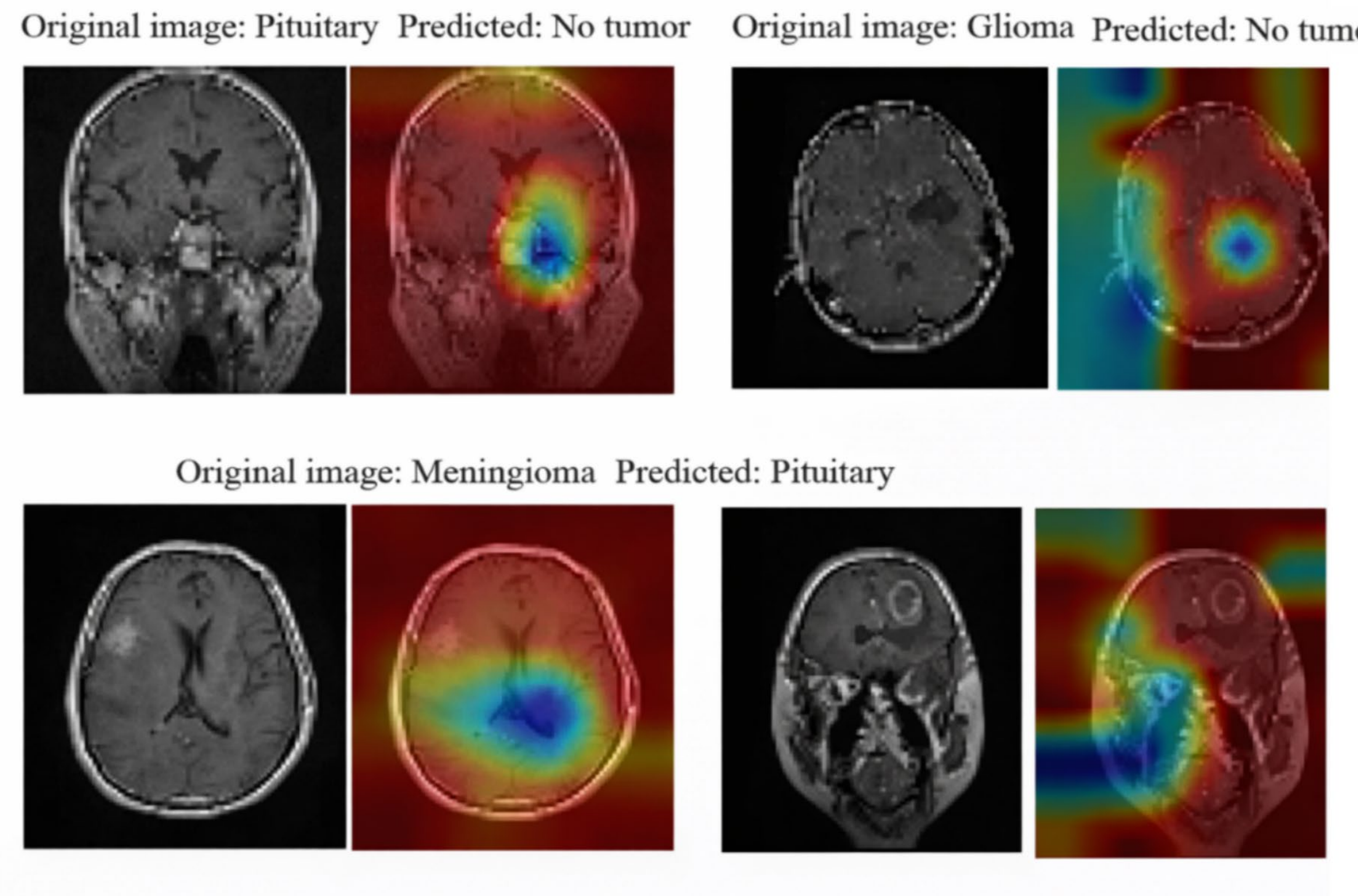

**Fig. 18** GRAD-CAM view of misclassified images

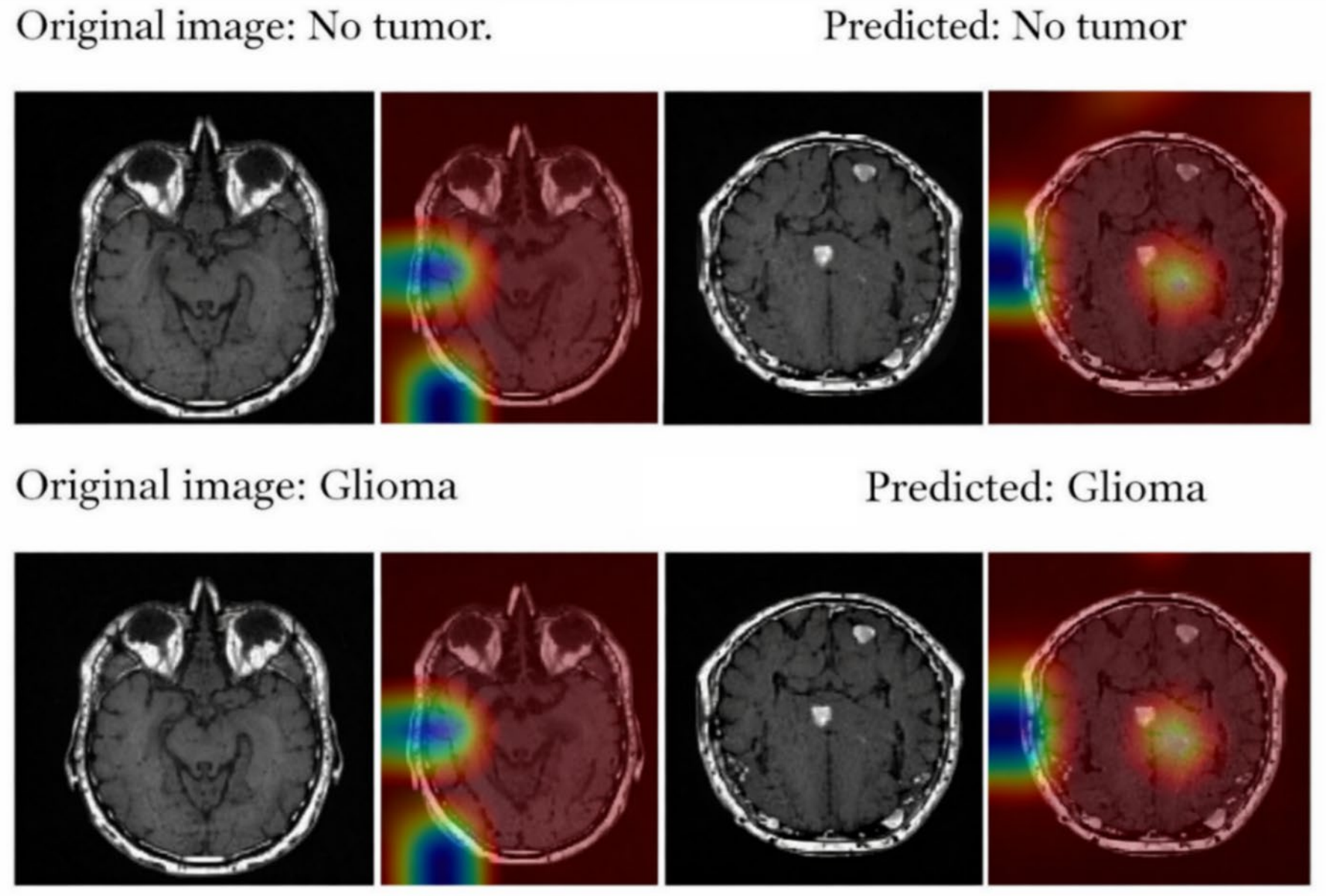

**Fig. 19** Grad-CAM view of correctly classified images

#### *4.7.4 Pixel intensity analysis of corrected/misclassified images*

This boxplot (see Fig. 22 and Table 15) compares the pixel-intensity distributions of misclassified and correctly classified instances. The correctly classified images (orange) have a higher mean pixel intensity (61.02) than the misclassified images (blue) (44.94). The boxplots and descriptive statistics suggest that images with higher pixel intensities are more likely to be correctly classified. Misclassified images exhibit lower variability (std = 29.81 vs. 36.99) and narrower intensity ranges, indicating that the model struggles when features are less prominent or less variable, leading to errors.

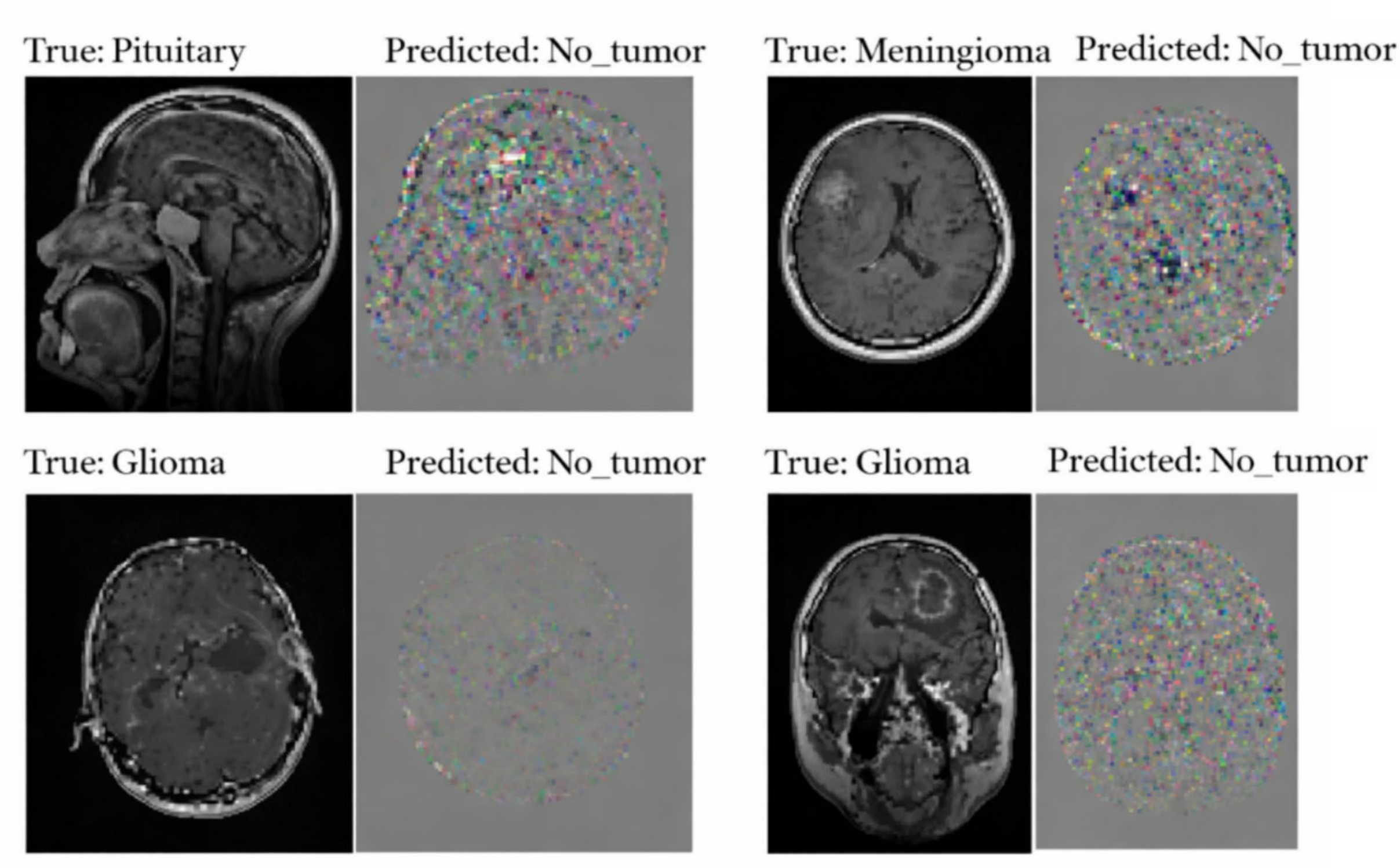

**Fig. 20** Grad-CAM view of the pixel intensity of misclassified images

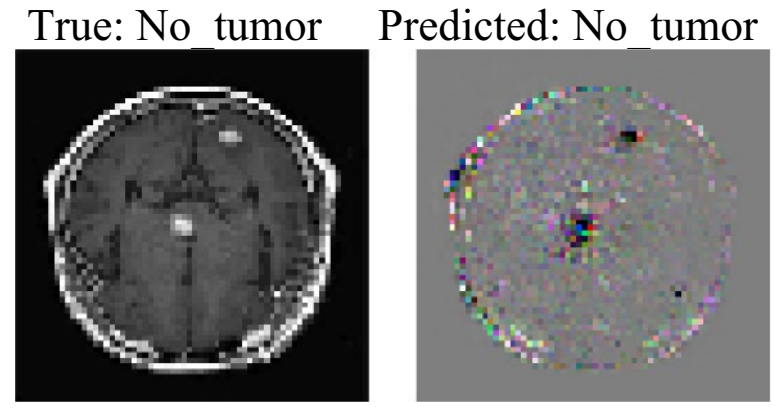

**Fig. 21** Grad-CAM view of the pixel intensity for correctly classified images

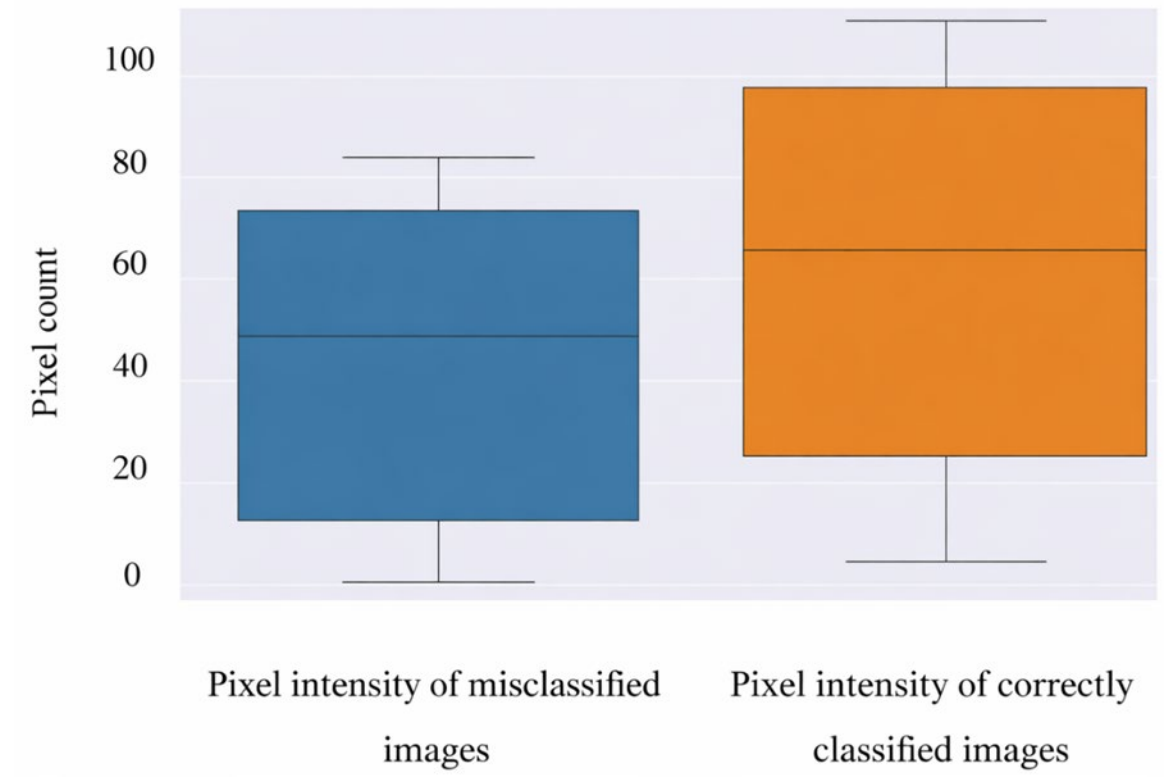

**Fig. 22** Distribution of pixel intensity of misclassified/correctly classified images

**Table 15** Descriptive statistics of the pixel intensity of correctly classified and misclassified samples

| Statistic | Misclassified | Correctly classified |
|---|---|---|
| Mean | 45 | 61 |
| Std | 30 | 36 |
| Min | 0.4 | 4 |
| 25% | 12 | 23 |
| 50% | 49 | 67 |
| 75% | 75 | 98 |
| Max | 85 | 113 |
| Interquartile range | 63 | 75 |

#### *4.7.5 Pixel intensity of TP-FP-TN-FN tumor modalities*

Figure 23 and Table 16 present the distributions of pixel intensities for true-positive, true-negative, false-positive, and false-negative cases. The *BrainFusionNet* performs best at mid-range pixel intensities, and errors occur at extreme or highly variable intensities (max = 168.75). The true positives (mean = 53.03, IQR = 57.14) and true negatives (mean = 47.66, IQR = 55.74) have more consistent distributions. The false positives (mean = 50.38, IQR = 47.58) have a narrower range and slightly lower

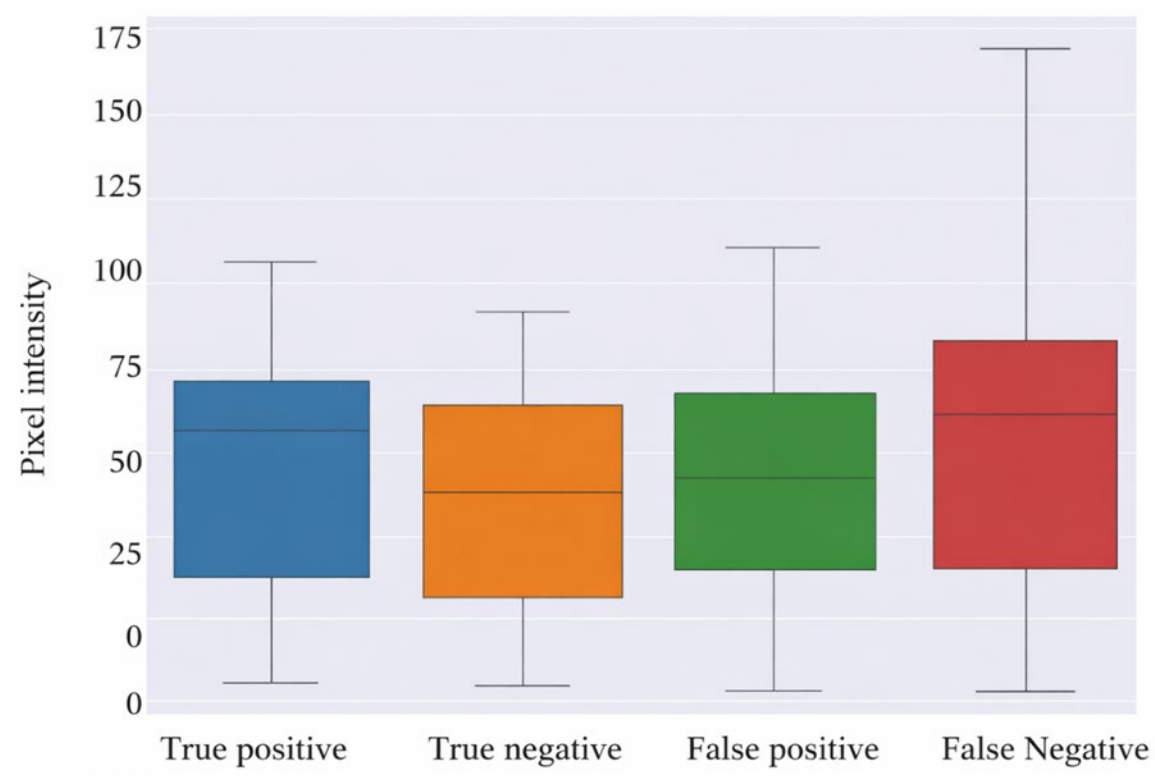

**Fig. 23** Distribution of the pixel intensity of TP-TN-FP-FN

**Table 16** Descriptive statistics for the pixel intensity of the TP-TN-FP-FN images

| Statistic | True positive | True negative | False positive | False negative |
|---|---|---|---|---|
| Mean | 53 | 48 | 50 | 62 |
| Std | 31 | 28 | 27 | 37 |
| Min | 2 | 2 | 4 | 1 |
| 25% | 22 | 20 | 27 | 30 |
| 50% | 65 | 49 | 52 | 69 |
| 75% | 79 | 76 | 75 | 87 |
| Max | 111 | 98 | 114 | 168 |
| Interquartile range | 57 | 56 | 47 | 56 |

variability, suggesting that these errors often occur in regions with less contrast.

#### *4.7.6 Evaluation of explanations*

To assess the correctness and usefulness of the generated explanations, we compare the highlighted regions in the LIME, SHAP, and Grad-CAM visualizations with the expert annotations from medical professionals. The correctness is measured by the alignment of these regions with known tumor areas and expected anatomical structures. We define "correct" explanations as those that highlight tumor regions consistent with expert knowledge and "useful" explanations as those that provide interpretable insights for medical practitioners to make more informed decisions. Additionally, consistency across methods (LIME, SHAP, Grad-CAM) serves as an indicator of reliability, with agreement among methods strengthening the validity of the explanation.

For instance, the LIME-generated heatmaps for Glioma (Fig. 16) indicate a substantial positive contribution from regions typically associated with glioma growth, such as the peritumoral zone, which is known for its distinct imaging signature. Similarly, SHAP visualizations (Fig. 17) highlight tumor regions with high contrast, matching the typical features of meningiomas and gliomas.

#### *4.7.7 Limitations of XAI methods*

While LIME, SHAP, and Grad-CAM provide valuable insights into model behavior, each has limitations. Grad-CAM, for instance, can sometimes focus on irrelevant regions of an image due to its reliance on gradients, leading to false positives in identifying tumor areas. LIME and SHAP, while more interpretable, may not always capture the nuanced tumor characteristics, particularly when tumors are small or have ambiguous boundaries. Additionally, these methods require significant computational resources, especially when working with large image datasets, which may limit their practical application in real-time clinical settings. These limitations highlight the need for further research to improve the spatial accuracy and computational efficiency of XAI techniques.

## 5 Discussion

In this study, a hybrid *BrainFusionNet* was proposed to improve the accuracy of brain tumor detection and classification. Our model provided better accuracy than Agarwal et al. [4], Ahmed et al. [6, ].

Our *BrainFusionNet* model, with fewer layers and 5,603,076 learnable parameters, reliably detects tumors in MRI images (see Fig. 24). Prior studies emphasized the need for DL on edge devices for real-time detection and monitoring of brain tumors. Figure 24 compares the parameter counts of SOTA CNNs and *BrainFusionNet*, demonstrating that *BrainFusionNet* is a suitable model for edge devices. This is because the model will require fewer computational resources than the CNN.

The XAI implementation is not limited to generating heatmaps using LIME, SHAP, and Grad-CAM. Instead, our XAI implementation is much more extensive than only generating heatmaps [6, 9]. We applied the Grad-CAM view of the pixel intensity. Furthermore, the pixel intensity was analyzed using the statistical methods of true classification, misclassification, true positive, false positive, true negative, and false negative. This study contributes to our understanding of the growing body of literature on the use of explainable AI for improving medical image analysis and diagnosis. It provides insights into the interpretability and transparency of CNNs for brain cancer modalities.

Among the SOTA CNNs, DenseNet201 and VGG16 achieved the best accuracy (96%) (see Table 5). InceptionV3 and Xception achieved 93% accuracy, performing well in most classes. ResNet50 achieved 92% accuracy, which demonstrates reliable performance. MobileNetV2 performed the least, with 77% accuracy. Table 5 shows the classification results of the SOTA CNNs. Among the SOTA CNNs, the DenseNet architecture is more efficient at detecting and classifying MRI images. Our finding is also supported by Verma and Singh [71] and Narasimha Raju et al. [55]. Narasimha Raju et al. [55] applied

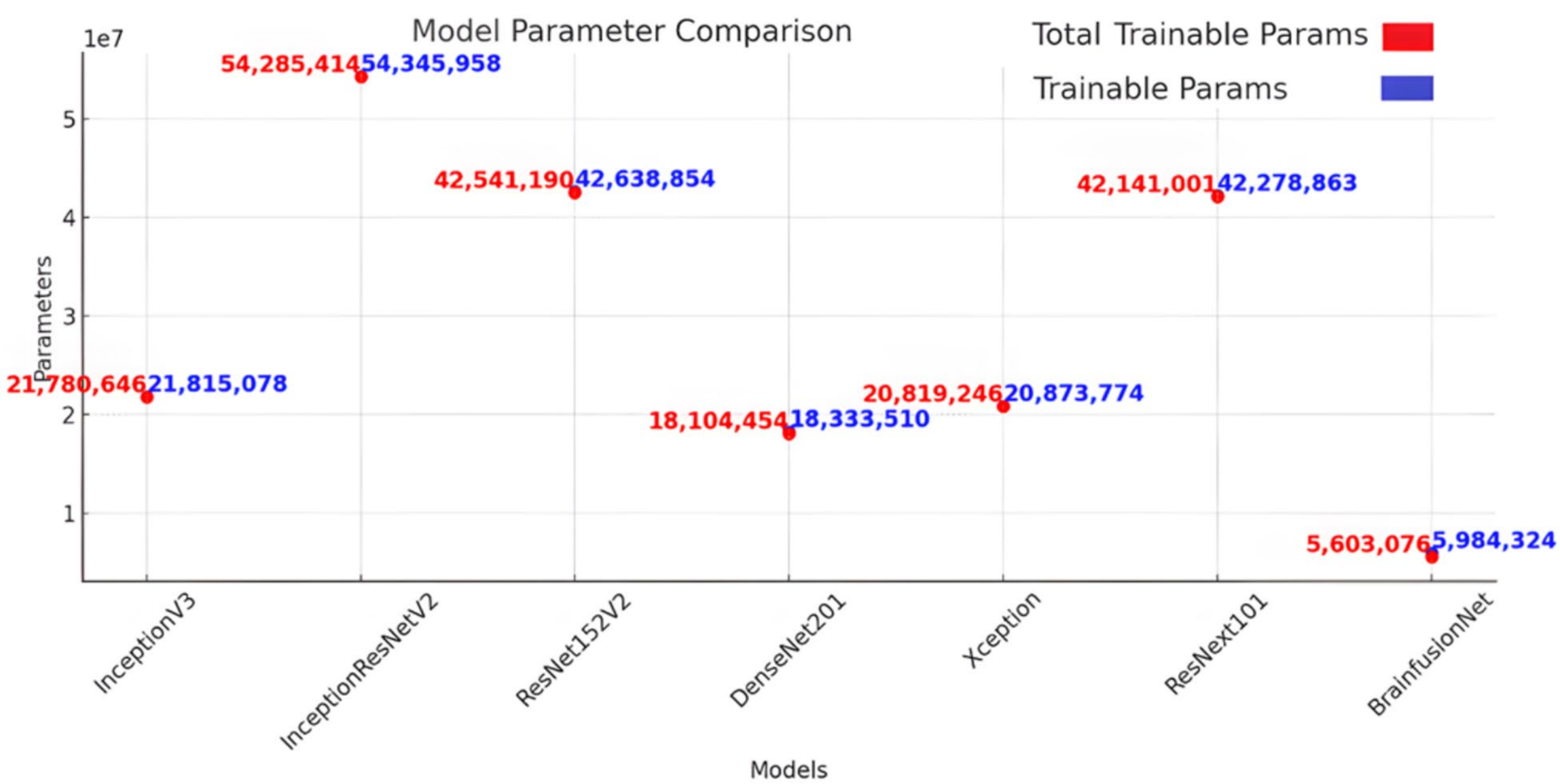

**Fig. 24** Comparison of the parameters of the SOTA CNNs and *BrainFusionNet*

DenseNet201 and achieved the best result (99.15%) for colorectal disease identification. Verma and Singh [71] reported that the DenseNet architecture uses densely flowing skip connections. The DenseNet connection pattern connects all layers directly, maximizing information flow within the network. The process ensures this. The feed-forward nature of each layer is maintained by receiving input from the previous layers. In medical datasets, class imbalance is also a common challenge for loss functions. However, certain cell types are underrepresented. In summary, DenseNet is a well-performing CNN model in which all the layers are connected to the network.

Many studies have shown that transfer learning improves classification accuracy and reduces training time compared with conventional CNNs [36, 37]. For example, Mujahid et al. [52] reported 95.07% accuracy in binary classification; Emam et al. (2024) reported breast cancer diagnosis using an optimized transfer learning technique, improving accuracy. However, in our study, transfer learning yielded negative results. The accuracy is lower than that of the main CNN. The findings of negative transfer learning support that, when the input image differs from the ImageNet-trained data, accuracy is likely to decrease. The effect of background noise and the application of different augmentation techniques, separately, on the test sets resulted in a decrease in performance. However, in the original CNN, the model was trained and tested on similar inputs, and its predictive performance improved on unseen data. Moreover, although CNNs can learn features regardless of the input data, this study's limited number of datasets is likely a factor limiting their predictive capability. Our view is also supported by Barbedo [15], who suggested that increasing the dataset size may improve transfer learning performance when the input images are augmented.

## 6 Limitations and Conclusions

The main limitation of this study is that the BrainFusionNet model was applied to the secondary dataset. In our defense, we would like to point out that biomedical image data collection is a difficult task, and the research was conducted in this context. Second, the model is applied only to MRI images. In the future, the model should be applied to computed tomography (CT), ultrasound (US), X-ray imaging (XR), Plasmid Bluescript (PBS), and microscopy images. Although transfer learning has negative results, future experiments should investigate how a hybrid CNN can leverage it.

This study presents a hybrid model, BrainFusionNet, that combines a CNN, ViT, and a GRU. The CNN was used to capture local features, and ViT was employed for global feature extraction. CNN and ViT identify crucial features in brain tumor MRI images, and the GRU establishes clear correlations among them. Compared with SOTA CNNs and transfer learning, the SMOTE model effectively addresses class imbalance in the MRI image dataset and achieves superior accuracy. The model was trained on two sets of brain tumor datasets obtained from Kaggle, achieving notable accuracies of 98% for 4-class and 3-class brain tumor classification. Six SOTA CNNs from different networks, including spatial exploitation (VGG19), depth-based (ResNet152v2), multipath (DenseNet201), width-based multiconnection (ResNext101), feature map exploitation (SE-ResNet152), and MobileNetV2, were tested on 7023 brain tumor MRI images. Second, transfer learning was applied.

The BrainFusionNet models, such as LIME, SHAP, and Grad-CAM, were integrated with XAI. This integration expands our understanding of how a model detects and classifies input images. Our experiment on pixel intensity in correctly classified and misclassified images also reveals that it impacts accuracy, as it directly influences correct and incorrect classifications. To our knowledge, such experiments in the field of brain tumors have rarely been conducted. Our rigorous experimental evaluation of models and pixel intensities provides a promising avenue for applying deep learning to the automated diagnosis of brain tumors and other cancer images.

Finally, since many scholars advocate integrating AI for practicality, we present a prototype that combines the BrainFusionNet model, a Streamlit-based web interface, and an Android mobile application. The prototype is expected to support ongoing efforts in the biomedical field to make CAD more practical, interpretable, and widely usable. Furthermore, while the core focus is on brain tumor classification, the broader framework has potential applications in other areas of disease detection, including agricultural disease management, where similar challenges around accessibility and decision support persist.

**Author contributions**

Md Taimur Ahad: Writing—original draft; Project administration; Methodology; Investigation; Formal analysis; Conceptualization. Dr. Bo Song: Writing—original draft; Writing—revision editing; Conceptualization; Formal analysis. Professor Yan Li: Project administration; Conceptualization.

**Funding**

Open Access funding enabled and organized by CAUL and its Member Institutions. This research did not receive any specific grant from funding agencies in the public, commercial, or not-for-profit sectors.

**Data availability**

The datasets used in this investigation are publicly available. The MRI dataset is accessible via https://www.kaggle.com/datasets/hey24sheep/rsnabreastcancerdataset. The links to the datasets are: https://www.kaggle.com/datasets/ashkhagan/figshare-brain-tumor-dataset https://www.kaggle.com/datasets/sartajbhuvaji/brain-tumor-classification-mri https://www.kaggle.com/code/ahmedhamada0/brain-tumor-detection-br35h

## Declarations

**Competing interest**

The authors declare no competing interest.

## Publisher's Note